\documentclass[10pt,twocolumn,letterpaper]{article}

\usepackage{iccv}
\usepackage{times}
\usepackage{epsfig}
\usepackage{graphicx}
\usepackage{amsmath}
\usepackage{amssymb}
\usepackage{caption}
\usepackage{subcaption}
\usepackage{makecell}
\usepackage[outercaption]{sidecap}
\usepackage{etoolbox}
\usepackage{multirow}
\usepackage{subfiles}

\usepackage[pagebackref=true,breaklinks=true,letterpaper=true,colorlinks,bookmarks=false]{hyperref}

\usepackage[accsupp]{axessibility}  

\newcommand{\MyMapTemplatePrefixc}[4]{\expandafter#1\csname#3#4\endcsname{#2{#4}}} 
\forcsvlist{\MyMapTemplatePrefixc {\def} {\mathcal}{c}} {A,B,C,D,E,F,G,H,I,J,K,L,M,N,O,P,Q,R,S,T,U,V,W,X,Y,Z}  

\newcommand{\MyMapTemplatePrefixtb}[5]{\expandafter#1\csname#4#5\endcsname{#2{#3{#5}}}} 
\forcsvlist{\MyMapTemplatePrefixtb {\def} {\tilde}{\mathbf}{t}} {A,B,C,D,E,F,G,H,I,J,K,L,M,N,O,P,Q,R,S,T,U,V,W,X,Y,Z}  
\forcsvlist{\MyMapTemplatePrefixtb {\def} {\tilde}{\mathbf}{t}} {0,1,a,b,c,d,e,f,g,h,i,j,k,l,m,n,o,p,q,r,s,u,v,w,x,y,z}  

\DeclareMathOperator*{\argmax}{arg\,max}

\newcommand{\MyMapTemplateNoPrefix}[3]{\expandafter#1\csname#3\endcsname{#2{#3}}}
\forcsvlist{\MyMapTemplateNoPrefix {\def} {\mathbf} } {0,1,a,b,c,d,e, f, g, h, i, j, k, l, m, n, o, p, q, r, u, v, w, x, y, z} 
\forcsvlist{\MyMapTemplateNoPrefix {\def} {\mathbf} } {A,B,C,D,E,F,G,H,I,J,K,L,M,N,O,P,Q,R,S,T,U,V,W,X,Y,Z}  

\def\ie{{\it i.e., }}

\iccvfinalcopy 

\ificcvfinal\fi

\begin{document}

\title{mDALU: Multi-Source Domain Adaptation and Label Unification with Partial Datasets}
\author{Rui Gong \textsuperscript{\rm 1}, Dengxin Dai \textsuperscript{\rm 1,4}, Yuhua Chen \textsuperscript{\rm 1}, Wen Li \textsuperscript{\rm 3}, Luc Van Gool \textsuperscript{\rm 1,2}\\
\textsuperscript{\rm 1} Computer Vision Lab, ETH Zurich, \textsuperscript{\rm 2} VISICS, KU Leuven, \textsuperscript{\rm 3} UESTC, \textsuperscript{\rm 4} MPI for Informatics\\
{\tt\small\{gongr, dai, yuhua.chen, vangool\}@vision.ee.ethz.ch, liwenbnu@gmail.com}
}

\maketitle
\ificcvfinal\thispagestyle{empty}\fi

\begin{abstract}
One challenge of object recognition is to generalize to new domains, to more classes and/or to new modalities. This necessitates methods to combine and reuse existing datasets that may belong to different domains, have partial annotations, and/or have different data modalities. This paper formulates this as a multi-source domain adaptation and label unification problem, and proposes a novel method for it. Our method consists of a partially-supervised adaptation stage and a fully-supervised adaptation stage. 
In the former, partial knowledge is transferred from multiple source domains to the target domain and fused therein. Negative transfer between unmatching label spaces is mitigated via three new modules: domain attention, uncertainty maximization and attention-guided adversarial alignment. In the latter,  knowledge is transferred in the unified label space after a label completion process with pseudo-labels. Extensive experiments on three different tasks - image classification, 2D semantic image segmentation, and joint 2D-3D semantic segmentation - show that our method outperforms all competing methods significantly.
\end{abstract}

\section{Introduction}
\label{sec:intro}

The development of object recognition is carried by two pillars: large-scale data annotation and deep neural networks. With new applications coming out every day, researchers need to constantly develop new methods and create new datasets. While we are able to develop novel neural networks for new tasks, the creation of new datasets can hardly keep up due to its huge cost. In the literature, a diverse set of learning paradigms, such as self-learning~\cite{he2019moco}, semi-supervised learning~\cite{hoyer2021three} and transfer learning~\cite{DomainAdaptiveFasterRCNN}, have been developed to come to the rescue. We enrich this repository by developing a method to combine multiple existing datasets that have been annotated in different domains, for smaller-scale tasks (fewer classes), and/or with fewer data modalities. The importance of the method can be justified by the fact that as time goes, research goals will become more and more ambitious, so object recognition models for more classes, new domains, and/or more data modalities are  necessary.

\begin{figure}
    \centering
    \includegraphics[width=0.9\linewidth]{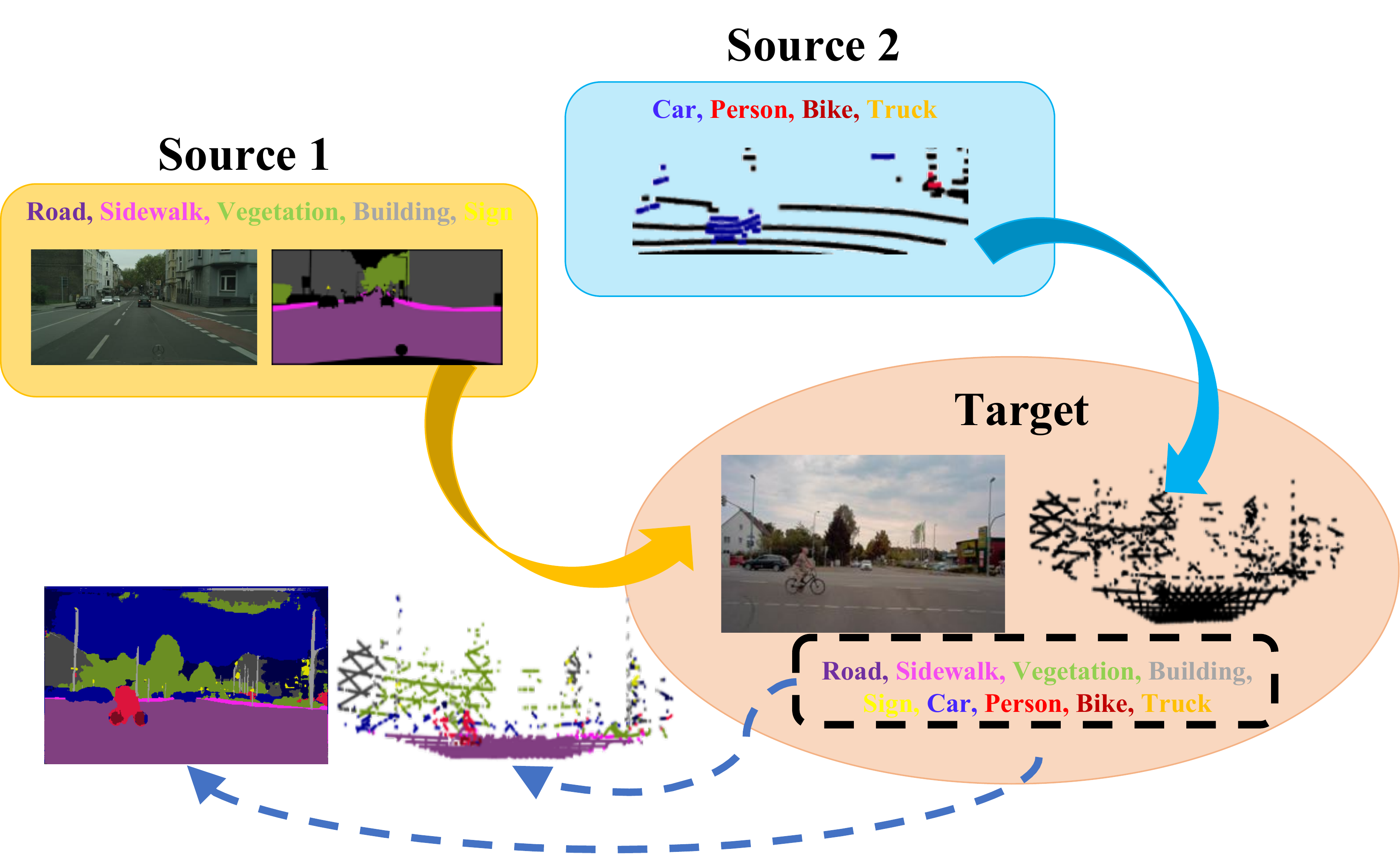}
    \caption{mDALU learns a complete-class and complete-modality object recognition model for a new, unlabeled target domain, by using multiple datasets with partial-class annotation and partial data modality as source domains.}
    \label{fig:mDALU}
    \vspace{-10pt}
\end{figure}

To address this, we propose a multi-source domain adaptation and label unification (mDALU) problem. In this setting, there are multiple source domains and an unlabeled target domain. In each source domain, only samples (images, pixels, or LiDAR points) belonging to a subset of classes are labeled; the rest are unlabeled. The subsets of classes having labels can be different over different source domains, and can have inconsistent taxonomies, \eg, truck is labeled as ``truck"  in one source domain but labeled as ``vehicle"  together with other types of vehicles in another. Further, the data modalities in different source domains can also be different, \eg, one contains images and the other contains LiDAR point clouds. The goal is to obtain an object recognition model for all classes in the target domain. Fig. \ref{fig:mDALU} shows an exemplar setting of mDALU. A comparison to other domain adaptation settings, in Table~\ref{tab:da:summary}, shows that mDALU is very flexible.

\begin{table*}[bt]
    \centering
    \resizebox{\textwidth}{12mm}{
    \begin{tabular}{c|c|c|c|c|c|c}
        \hline
        
        \hline
         \makecell{Domain Adaptation Setting} & \makecell{Can Handle Multiple \\ Source Domains?} & \makecell{Can Handle Multiple \\ Data Modalities?} &  \makecell{Can Handle Different Label \\ Spaces of Source Domains?} &  \makecell{Change of Label Space Size\\ from Source to Target Domain} & \makecell{Can Handle Partial \\ Annotations?} & \makecell{Can Handle Inconsistent \\ Taxonomy?}\\
        \hline
        Unsupervised Domain Adaptation \cite{advesarial:alignment:15} & No & No & $-$ & Same Size & No & $-$\\
        Partial Domain Adaptation \cite{partial:domain:eccv18} &  No &  No & $-$ & Reduced & No & $-$\\
        Multi-Source Domain Adaptation \cite{moment:matching:DA:iccv19, zhao2019multi} & Yes  &  No & No &  Same Size & No & No\\
        Category-Shift Multi-Source Domain Adaptation \cite{xu2018deep} & Yes & No & Yes & Increased & No & No\\
        Multi-Modal Domain Adaptation \cite{Jaritz_2020_CVPR} & Yes & Yes & No & Same Size & No & No\\
        Multi-Source Open-Set Domain Adaptation \cite{openset:DA:eccv20,openset:da:iccv17} & Yes & No & No & Same Size + 1$^*$ & Yes & No \\
        \hline
        \hline
        Multi-Source Domain Adaptation and Label Unification (mDALU) & \textbf{Yes} & \textbf{Yes} & \textbf{Yes} & \textbf{Increased} & \textbf{Yes} & \textbf{Yes}\\
        \hline
        
        \hline
    \end{tabular}
    }
    \caption{Comparison between our mDALU and other domain adaptation settings (see Sec.~\ref{sec:related} for details). It is clear that mDALU offers a very flexible and general setting. $^*$ ``1" means an additional ``unknown" class in the target domain.}
    \label{tab:da:summary}
    \vspace{-10pt}
\end{table*}

This goal is challenging. Firstly, there is the notorious issue of negative transfer. While negative transfer is an issue also for standard transfer and multi-task learning, it is especially severe in our mDALU task due to the influence of unlabeled classes. To address this, we propose three novel modules, termed domain attention, uncertainty maximization and attention-guided adversarial alignment, to avoid making confident predictions for unlabeled samples in the source domains, and to enable robust distribution alignment between the source domains and the target domain. The method with the aforementioned modules and attention-guided prediction fusion is able to generate good results in the unified label space and on the target domain. In order to further improve the results, we need to fuse the supervision of all partial datasets to transfer the supervision in the unified label space. To this aim, we propose a pseudo-label based supervision fusion module. In particular, we generate pseudo-labels for the unlabeled samples in the source domains and all samples in the target domain. Standard supervised learning is then performed in the unified label space for the final model. 

To showcase the effectiveness of our method, we evaluate it on three different tasks: image classification, 2D semantic image segmentation, and joint 2D-3D semantic segmentation. Synthetic and real data, and images and LiDAR point clouds are involved. Also, non-overlapping, partially-overlapping and fully-overlapping label spaces, and consistent and inconsistent taxonomies across source domains are covered. Experiments show that our method outperforms all competing methods significantly.

\section{Related Work}
\label{sec:related}
\noindent
\textbf{Multi-Source Domain Adaptation}. Transfer learning and domain adaptation have been extensively studied in the past years. 
Several effective strategies have been developed such as minimizing maximum mean discrepancy~\cite{mmd:da:14,long2015learning}, moment matching~\cite{zellinger2017central}, adversarial domain confusion~\cite{advesarial:alignment:15,tzeng2017adversarial}, entropy regularization~\cite{vu2018advent}, and curriculum domain adaptation~\cite{SynRealDataFog19}. 
While great progress has been achieved, most algorithms focus on the single-source adaptation setting. This limits the methods from being used when data is collected from multiple source domains. 
That is why multi-source domain adaptation methods are proposed~\cite{JMLR:v9:crammer08a,advesarial:MSDA:nips18,moment:matching:DA:iccv19,msda:alg:theory:18,zhao2019multi}. Yet, these methods all assume the same label space for all domains. Xu \etal~\cite{xu2018deep} explores the problem of the category shift among different source domains, and adopts the k-way domain discriminator to reduce the effect of category shift. But the method is mainly proposed for the image classification task, and cannot deal with the problem of partial annotation, inconsistent taxonomies and modal differences among different source domains. 

\noindent\textbf{Open-Set/Partial Domain Adaptation.} 
Recent research explores the category ``openness" between the source domain and the target domain, which is divided into open-set domain adaptation and partial domain adaptation. Open-set domain adaptation~\cite{openset:da:iccv17,openset:DA:eccv18,openset:DA:eccv20} assumes that the target domain includes new classes that are unseen in the source domain, and aims to classify the unseen class samples as ``unknown" class in the target domain. Partial domain adaptation \cite{Cao_2018_CVPR,partial:domain2:cvpr18,partial:domain:eccv18,partial:adversarial:20} aims to transfer knowledge from existing large-scale domains (e.g. 1K classes) to unknown small-scale domains (e.g. 20 classes) for customized applications. Different than both open-set and partial domain adaptation, our label space of the target domain is the union of label spaces of all source domains.

\noindent
\textbf{Learning from multiple datasets}. 
Several successful methods~\cite{residual:adapter:nips17,rebuffi-cvpr2018,wang2019towards,universal:ssl:segmetnation:iccv19} have been proposed to learn a single universal network, that can represent different domains with a minimum of domain-specific parameters. 
But those methods do not consider domain adaptation and label space unification. 
Recently, Lambert \etal \cite{MSeg:composite:dataset:cvpr20} presented a composite dataset that unifies different semantic segmentation datasets by reconciling the taxonomies, merging and splitting classes manually. But they do not address the problem of domain adaptation, partial annotation and cross-modal data, and they rely on the manual re-annotation for unification.
The  object detection method by Zhao \etal~\cite{Zhao_UniDet_ECCV20} performs label space unification from multiple datasets with partial annotations, but it does not consider other problems that are considered by our method such as domain discrepancies, inconsistent taxonomies and mismatched data modalities across the datasets.

\section{Approach} 
\label{sec:approach}
\subsection{Problem Statement}
For the problem of mDALU, we are given $K$ source domains $\cS_{1}, \cS_{2}, ..., \cS_{K}$. The $K$ source domains contain the samples from $K$ different distributions $P_{S_{1}}, P_{S_{2}}, ..., P_{S_{K}}$, which are labeled with $C_{1}, C_{2}, ..., C_{K}$ classes, resp. All the source domains can contain both partially labeled and unlabeled samples. The unlabeled samples can belong to the labeled classes of other domains. The label spaces $\cC_{1}, \cC_{2}, ..., \cC_{K}$ can be non-, partially-, or fully-overlapping with each other. Moreover, both consistent and inconsistent taxonomies among $\cC_{1}, \cC_{2}, ..., \cC_{K}$ are allowed. Then the union of the label spaces $\cC_{i}, i=1, ..., K$ forms the unified and complete label space $\cC_{\cup} = {\cC_{1}\cup \cC_{2}\cdots\cC_{K}}$, including $C_{\cup}$ classes. Besides, the unlabeled target domain $\cT$ is given, containing samples from the distribution $P_{T}$. Denoting the source samples $\x^{s_{i}}\in \cS_{i}, i=1,...,K$ and the target samples $\x^{t}\in \cT$, we have $\x^{s_{i}}\sim P_{S_{i}}, \x^{t}\sim P_{T}$, $P_{S_{1}}\neq P_{S_{2}}\neq ...\neq P_{S_{K}} \neq P_{T}$. The mDALU problem aims at training the model on the $K$ source domains $\cS_{i}, i=1, ...,K$, labeled with $C_{i}$ classes in each, and the unlabeled target domain $\cT$, to improve the performance of the model on the target domain $\cT$ in the unified label space $\cC_{\cup}$. We use $\y^{s_i}$ to indicate the ground-truth label map of $\x^{s_{i}}$. Note that we present most of our approach with the notation of 2D semantic image segmentation. The translation to image classification and 3D point cloud segmentation is straightforward -- by replacing pixels with images and by replacing pixels with 3D LiDAR points. 

\subsection{Our Approach to mDALU problem}

\begin{figure}
    \centering
    \begin{subfigure}[b]{0.4\textwidth}
    \centering
    \includegraphics[width=\textwidth]{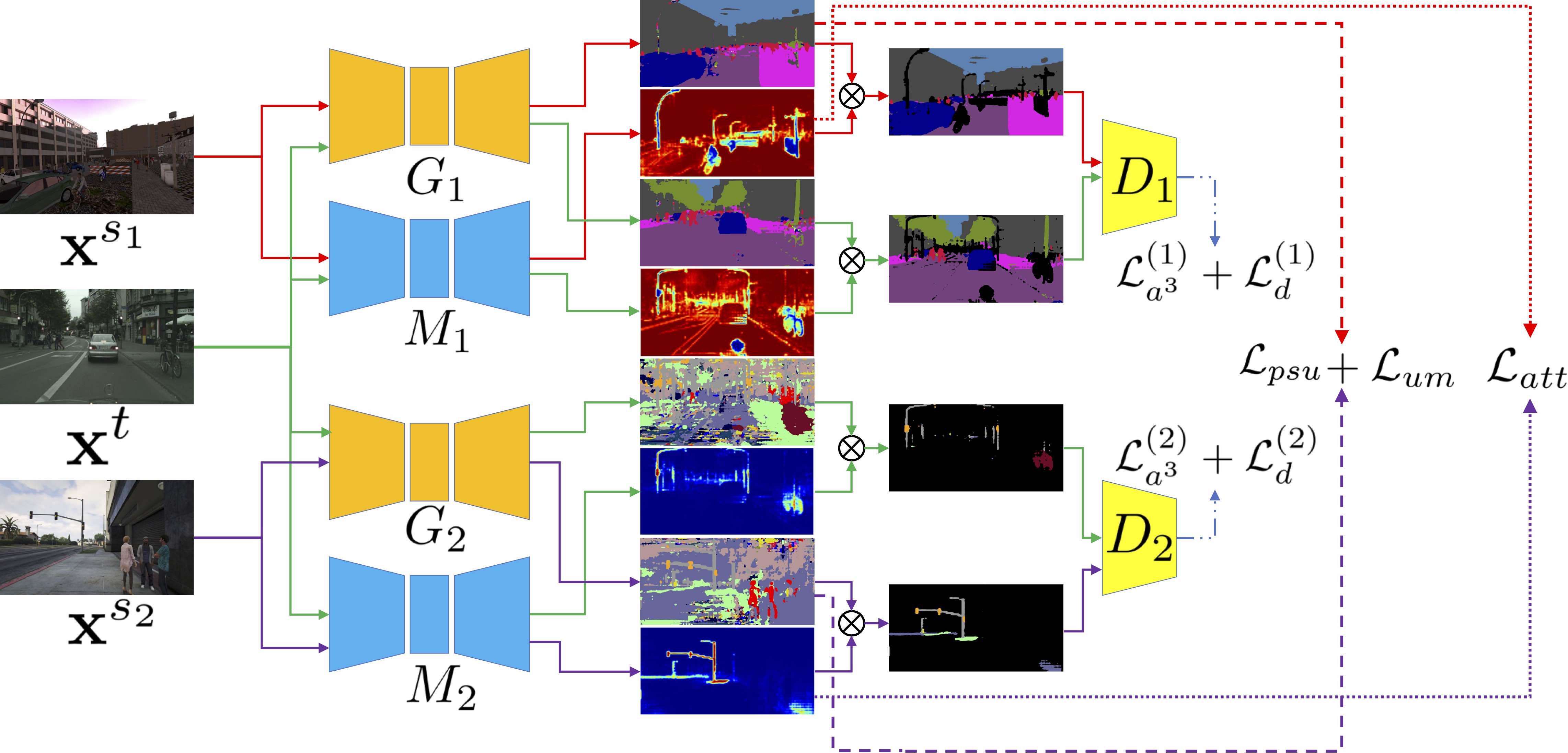}
    \caption{Partially-Supervised Adaptation}
    \end{subfigure}
    \begin{subfigure}[b]{0.4\textwidth}
    \centering
    \includegraphics[width=\textwidth]{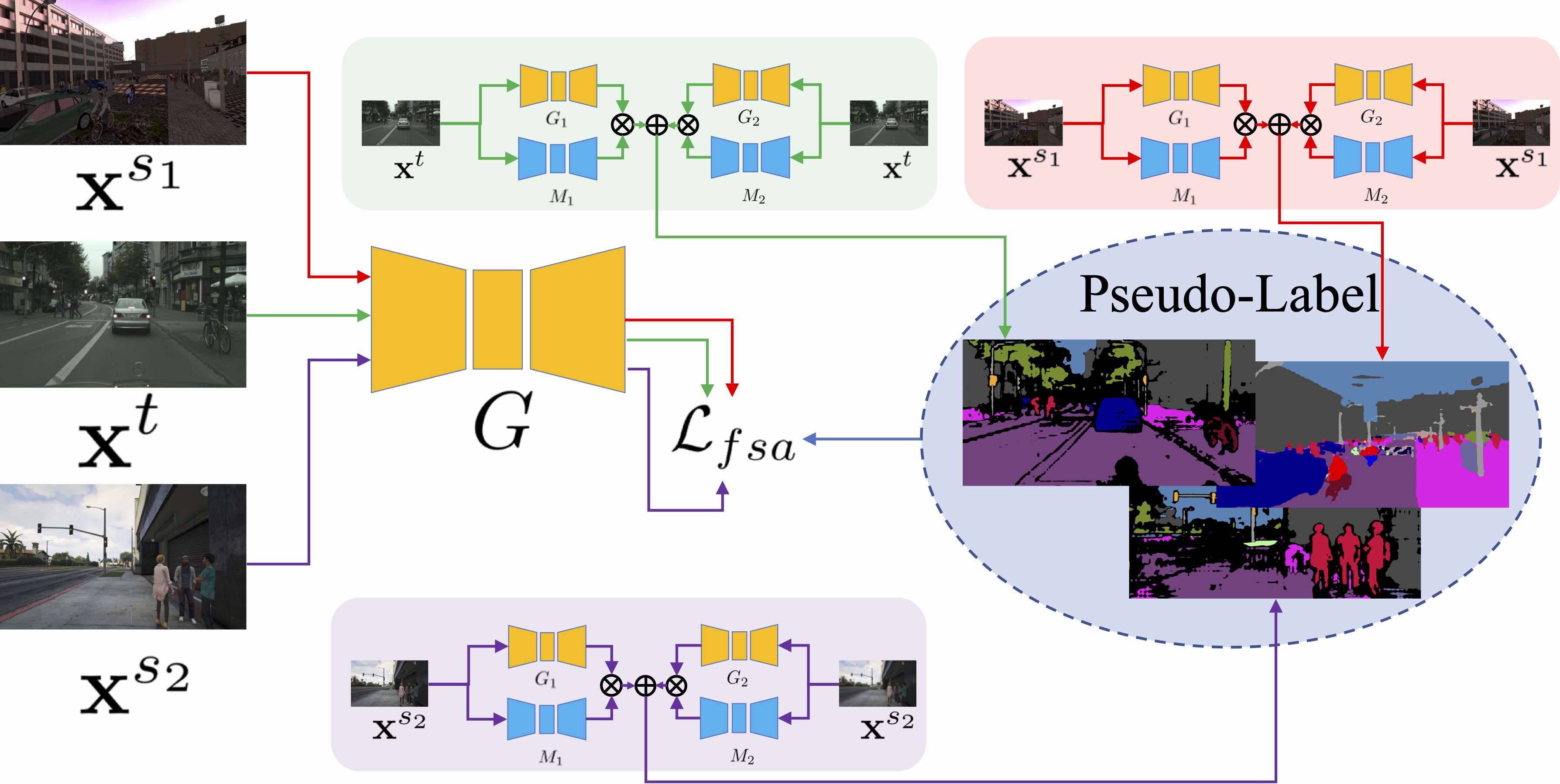}
    \caption{Fully-Supervised Adaptation}
    \end{subfigure}
    \vspace*{-10pt}
    \caption{Illustration of our approach to mDALU. There are 2 stages, (a) partially supervised adaptation and (b) fully-supervised adaptation.}
    \label{fig:approach}
    \vspace{-10pt}
\end{figure}

As shown in Fig.~\ref{fig:approach}, there are two stages in our approach, the partially-supervised adaptation stage and the fully supervised adaptation stage.
In the partially-supervised adaptation stage, the partial supervision is transferred to the target domain from different source domains, respectively. Then in the fully-supervised adaptation stage, the supervision, in complete label space, is fused and self-completed on the unlabeled samples, and jointly transferred in the source domains and target domain. In order to realize adaptation under partial supervision, we propose three modules: DAT, UM and A$^3$ for the first stage. Then in the second stage, we use PSF and further learning. Below we provide details of all these components. 
From Sec. \ref{subsec:psl} to Sec. \ref{subsec:psf}, we first introduce our method for mDALU under  consistent taxonomies. In this part, we first describe a basic version of our method composed of DAT and inference via attention-guided fusion, which will be followed by UM and A$^3$ to enhance the adaptation ability. Finally, we present PSF. Then in Sec. \ref{sec:inconsistent_taxonomy}, we extend our proposed method towards mDALU under inconsistent taxonomies.

\vspace{-5pt}\subsubsection{Partially-Supervised Learning}\label{subsec:psl}

Different segmentation networks $G_{i}, i=1,...,K$ are adopted for different source domains $\cS_{i}$. While their annotations cover partial label spaces $\mathcal{C}_i$, we train each network $G_{i}$ in the unified label space $\cC_{\cup}$ -- some classes have no training data -- with a standard cross-entropy loss $\cL_\text{psu}$. 
The network $G_{i}$ is composed of a feature extractor $E_i$ and a label predictor $B_{i}$, \ie $G_{i} = \{E_i, B_i\}$. While we can average the results of these models directly in the target domain for predictions in the unified label space, coined multi-branch (MBR) fusion, this generates poor results. This is because the predictions of each model $G_{i}$ for its unlabeled classes in $\cC_{\cup} \setminus \cC_i$ can be arbitrary numbers that dominate the averages. We thus propose the domain attention (DAT) module, which learns the attention map for $G_{i}$ to signal on which area its prediction is reliable, for more effective fusion.  

The attention map $\a^{s_i}$ in domain $\cS_i$ is defined as: 
\begin{eqnarray}
    \a^{s_i}(h,w)
    \begin{cases}
    = 1, \text{if } \y^{s_i}(h,w)\in\mathcal{C}_{i}\\
    = 0, \text{if }  \y^{s_i}(h,w)=\texttt{void},
    \end{cases}
\end{eqnarray}
where $(h,w)$ are pixel indices and \texttt{void} means no label. We train an attention network $M_i$ for each source domain $\cS_i$. The attention maps are predicted as $\tilde{\a}^{s_i} = M_i(\x^{s_i})$ and $\tilde{\a}_{i}^{t} = M_i(\x^{t})$. The attention network $M_i$ is composed of the feature extractor $E_i$ and a new label predictor $B_i^M$: $M_i=\{E_i, B_i^M\}$. $M_i$ is trained under an MSE loss $\cL_\text{att}$, together with $G_{i}$ in a multi-task setting.

\vspace{-5pt}\subsubsection{Inference via Attention-Guided Fusion}\label{sec:attfuse}
We feed an image $\x$ into semantic segmentation networks $G_i$ to generate the corresponding probability maps  $\hat{\p}_i\in [0,1]^{H\times W \times C_{\cup}}$, and into different attention networks $M_i$ to generate attention maps $\hat{\a}_{i}$. Then we fuse the predictions by averaging $\hat{\p}_i$ weighted by $\hat{\a}_{i}$:
\begin{eqnarray}
\label{eq:fusion}
    \f = \frac{\sum_{i=1}^{K}\hat{\a}_{i}\otimes \hat{\p}_i}{\sum_{j=1}^{C_{\cup}}(\sum_{i=1}^{K}\hat{\a}_{i}\otimes \hat{\p}_i)^{(j)}}, 
\end{eqnarray}
where $(\sum_{i=1}^{K}\hat{\a}_{i}\otimes \hat{\p_i})^{(j)}$ yields the probability of the $j^{\text{th}}$ class. The predicted class is then obtained via \texttt{argmax}.

\subsubsection{Uncertainty Maximization (UM)}
Due to the lack of ground truth class supervision, while we have the attention-guided fusion, the wrong prediction of unlabeled samples in the source domains can still have negative effects for our cross-domain prediction fusion. In order to further reduce the negative effects of unlabeled samples $\x_u^{s_i}$ in source domains, we propose a module specifically to maximize uncertainties of the predictions on unlabeled samples in those domains. In particular, $G_{i}(\x_u^{s_i})$ is expected to equally spread the probability mass to all classes, \ie obeying the uniform categorical distribution $\cU(C_{\cup})$. The probability density function $q(j)$ of $\cU(C_{\cup})$ is formulated as $q(j) = \frac{1}{C_{\cup}}$,
where $j=1,2,...,C_{\cup}$ is to represent different classes.  The probability distribution of the network prediction on unlabeled samples $G_i(\x_u^{s_i})$ is denoted as
$p(j)=G_i(\x_u^{s_i})^{(j)}$,
where $G_i(\x_u^{s_i})^{(j)}$ represents the probability of the $j^{\text{th}}$ class. In order to maximize the uncertainty of the prediction on the unlabeled samples, the distribution distance between $p(j)$ and $q(j)$ is expected to be minimized. Following the distribution distance metric in \cite{chen2019domain:cvpr2019}, we adopt the Pearson $\chi^2$-divergence for measuring the distribution distance, which is formulated as,
\begin{eqnarray}
    D_{\chi^2}(p||q) = \int_j ((\frac{p(j)}{q(j)})^2-1)q(j) ,\\
    D_{\chi^2}(p||q) = C_{\cup}\sum_{j=1}^{C_{\cup}}p(j)^2 - 1 .\label{eq:diveq} 
\end{eqnarray}
On the basis of Eq. (\ref{eq:diveq}), we propose the square loss $\cL_{um}$ for minimizing the Pearson $\chi^2$-divergence, \ie maximizing the uncertainty of the prediction on the unlabeled samples. $\cL_{um}$ can be written as
\begin{eqnarray}
    \cL_{um} = \sum_{j=1}^{C_{\cup}}(G_{i}(\x_u^{s_i})^{(j)})^2.
\end{eqnarray}
Through the UM module, we encourage the model to make uniform categorical probability predictions, $\frac{1}{C_{\cup}}$, for unlabeled samples over the unlabeled classes, to best preserve the uncertainty to let the ground truth supervision of those classes from other source domains make the decision in the further attention-guided fusion and PSF process.

\vspace{-10pt}\subsubsection{Attention-Guided Adversarial Alignment (A$^3$)}
It has been proven in the literature that adversarial alignment is effective for domain adaptation. We extend the idea to mDALU. For adversarial alignment, one discriminator $D_i$ is used for each source domain, to align the distribution between the source domain $\cS_i$ and the target domain $\cT$.  In general unsupervised domain adaptation, the discriminator training loss $\cL_d$ and the adversarial loss $\cL_{adv}$ \cite{Tsai_adaptseg_2018} for the source domain $\cS_i$ and the target domain $\cT$ is defined as 
\begin{eqnarray}
    \cL_{adv}^{(i)}(\x^t) = -\log(D_i(G_i(\x^t))) \label{eq:adv_ge}\\
    \cL_{d}^{(i)}(\x_i^{s_i}, \x^t) = -\log(D_i(G_i(\x^{s_i}))) \nonumber \\
    -\log(1-D_i(G_i(\x^{t}))). \label{eq:dis_gen}
\end{eqnarray}
 
 Yet, in our mDALU problem, there is no ground truth label guidance available for the unlabeled classes. A direct alignment between the source domain and the target domain will cause negative transfer, \ie the transfer of incorrect knowledge from the unlabeled parts in the source domains to the target domain. 
 Here, we again use our attention map $\a^{s_i}$ to alleviate this problem by proposing an attention-guided adversarial loss:
\begin{eqnarray}
    \cL_{a^3}^{(i)}(\x^t) = -\log(D_i(G_i(\x^t)\otimes M_i(\x^t))),\\
    \cL_{d}^{(i)}(\x_i^{s_i}, \x^t) = -\log(D_i(G_i(\x^{s_i})\otimes M_i(\x^{s_i}))) \nonumber \\
    -\log(1-D_i(G_i(\x^{t})\otimes M_i(\x^t))),
\end{eqnarray}
where $\otimes$ represents element-wise multiplication. 

Then the overall loss for our method at the first stage is:
\begin{eqnarray}
    \cL_{all} = \cL_{psu} +  \cL_{att} +  \cL_{um} + \lambda \sum_{i=1}^{K}\cL_{a^3}^{(i)},
    \label{eq:padv_detail}
\end{eqnarray}
where $\lambda$ is the hyper-parameter to balance out the attention-guided adversarial loss against other losses.
The whole optimization objective for our first partially-supervised domain adaptation stage can be formulated as:
\begin{eqnarray}
    \min_{G_i}\max_{D_i}  \cL_{all}.
\end{eqnarray}

\vspace{-5pt}\subsubsection{Pseudo-Label Based Supervision Fusion (PSF)} \label{subsec:psf}
In the first partially-supervised adaptation stage, knowledge in different label spaces $\cC_i$ is transferred from different source domains to the target domain. In the second fully-supervised adaptation stage, we aim at learning and transferring knowledge in the complete and unified label space $\cC_{\cup}$ between all domains jointly. In order to realize that, we  complete the label spaces for all the related domains $\cS_{1}, \cS_{2}, ..., \cS_{K}, \cT$ with pseudo-labels, \ie fuse the supervision from different label spaces $\cC_i$ to get the complete and unified supervision $\cC_{\cup}$. Here we present our pseudo-label based supervision fusion (PSF) method.

In order to complete the label space in the source domain $\cS_i$, we feed each of the source image samples $\x^{s_i}$ into every semantic model $G_k, k=1,...,K$, to generate `partial' semantic probability maps $\hat{\p}_k^{s_i}\in [0,1]^{H\times W \times C_{\cup}}$ and to every attention network $M_k, k=1,...,K$ for the attention map $\hat{\a}_k^{s_i}\in [0,1]^{H\times W}$. The fused prediction $\f^{s_i}$ is obtained via Eq.(~\ref{eq:fusion}).
We denote the predicted label map as $\bar{\y}^{s_i}$, generated by using an \texttt{argmax} operation over $\f^{s_i}$. 
The `pseudo-label' map $\hat{\y}^{s_i}$ for the source domain $\cS_i$ is defined as: 
\begin{equation}
\label{eq:psf}
\begin{split}
    \hat{\y}^{s_i}(h,w) =
  \begin{cases}
    \y^{s_i}(h,w), \text{if } \y^{s_i}(h,w) \neq \texttt{void}  \\
     \bar{\y}^{s_i}(h,w) \text{if } \y^{s_i}(h,w)=\texttt{void} \\ \text{ and } \f^{s_i}(h,w,\bar{\y}^{s_i}(h,w))>\delta
     \\
    \texttt{void}, \text{ otherwise} \\
    \end{cases}
    \end{split}
\end{equation}
where $\delta$ is a threshold determining whether to select the predicted pseudo-label.

On the target domain $\cT$, since no ground truth labels are available, we obtain pseudo labels directly from the predicted label map $\bar{\y}^{t}$ (obtained from $\f^{t}$ via an \texttt{argmax}):
\begin{equation}
    \hat{\y}^{t}(h,w) =
    \bar{\y}^{t}(h,w) \text{ if } \f^{t}(h,w,\bar{\y}^{t}(h,w))>\delta .
\end{equation}

By using the generated fused pseudo-label $\hat{\y}^{s_i}, \hat{\y}^{t}, i=1,...,K$, we complete the label space from $\cC_i$ to $\cC_{\cup}$ for the source domain $\cS_i$, and from $\emptyset$ to $\cC_{\cup}$ for the target domain $\cT$. We then train the network $G$ for all the related domains $\cS_1, \cS_2, ..., \cS_K, \cT$ with all the datasets in the unified label space. In total, the loss $\cL_{fsa}$ for our second `fully-supervised' adaptation stage is:
\begin{eqnarray}
    \cL_{fsa} = \sum_{i=1}^{K}  \cL_{ce}^{s_i} +  \cL_{ce}^{t},
\end{eqnarray}
where $\cL_{ce}$ is the standard cross-entropy loss.

\vspace{-5pt}\subsubsection{Inconsistent Taxonomies} \label{sec:inconsistent_taxonomy}
The above method is able to deal with the mDALU problem under consistent taxonomies, \ie the different classes in all source domains are exclusive with each other. Yet, there might be inconsistent taxonomies between different source domains, causing a performance drop for the inconsistent taxonomies classes. Here, we introduce the extension of our above method, to handle the inconsistent taxonomies problem.
Denoting the classes in the label spaces $\cC_i$ as $\c_i^o$, we have $\cC_i =\{ \c_i^o, o=1,2,..., C_i\}$. Then the inconsistent taxonomies among different source domains can be defined as, $\exists \c_p^q \in \cC_{p}, \c_m^n \in \cC_{m}, p, m=1, ..., K, p\neq m, q=1, ..., C_{p}, n=1, ..., C_{m}$, we have $\c_p^q \neq \c_m^n, \text{and } \c_p^q \cap \c_m^n \neq \emptyset .$
The inconsistent taxonomies classes between different source domains $\cS_{p}$ and $\cS_{m}$ are denoted as $\c_p^q \in \cC_p$ and $\c_m^n \in \cC_m$. For example, the truck is labeled as ``truck" class $\c_p^q$ in one dataset $\cS_{p}$, while it is labeled as ``vehicle" class $\c_m^n$ together with other vehicles in another dataset $\cS_{m}$. Another typical example is  motorcycles being labeled as ``cycle" class $\c_p^q$ together with other cycles in one dataset $\cS_{p}$, but being labeled as ``vehicle" class $\c_m^n$ together with other vehicles in another dataset $\cS_{m}$. 
In the unified label space of the target domain, the conflict part $\c_p^q \cap \c_m^n$ is assigned to either $\c_p^q$ or $\c_m^n$ exclusively. Without loss of generality and for reasons of clarity, it is assumed that the $\c_p^q \cap \c_m^n$ is assigned to $\c_p^q$. Then in order to solve the conflict of $\c_p^q$ and $\c_m^n$, in the attention-guided fusion, we introduce the additional class-wise weight map $\w_i\in \mathbb{R}^{H\times W\times C_{\cup}}$, and Eq. (\ref{eq:fusion}) is extended to Eq. (\ref{eq:fusion_afterweight}),
\begin{equation}
    \w_i(h,w,j) = 
    \begin{cases}
    = v, \text{if} \argmax \hat{\p}_{i}(h,w) = q^{\prime} , \text{and  } i = p,\\
    \quad\!\!\!\!\text{and  } \argmax \hat{\p}_{m}(h,w) = n^{\prime}, \text{and  } j = q^{\prime}\\
    = 1, \text{otherwise}
    \end{cases}
    \label{eq:weight_map}
    \hspace{-15pt}
\end{equation}
\vspace{-13pt}
\begin{eqnarray}
    \f = \frac{\sum_{i=1}^{K}\hat{\a}_{i}\otimes \hat{\p}_i \otimes \w_i}{\sum_{j=1}^{C_{\cup}}(\sum_{i=1}^{K}\hat{\a}_{i}\otimes \hat{\p}_i\otimes \w_i)^{(j)}}, \label{eq:fusion_afterweight}
\end{eqnarray}
where $v>1$ in Eq.~(\ref{eq:weight_map}) is a hyper-parameter, set to 5.0. $v$ is used to increase the weight of class $\c_p^q$ of the corresponding prediction $\hat{\p}_p$ in Eq.~(\ref{eq:fusion_afterweight}), to convert $\c_p^q \cap \c_m^n$ to $\c_p^q$ in the prediction fusion. $q^{\prime}$, $n^{\prime}$ are the class indices of $\c_p^q$ and $\c_m^n$ in the unified label space $\cC_{\cup}$. Correspondingly, under inconsistent taxonomies, besides the unlabeled samples in the source domains being completed with the predicted pseudo-label as in Eq. (\ref{eq:psf}), the conflict part $\c_p^q \cap \c_m^n$, which is labeled as $\c_m^n$ originally in $\cS_{m}$, is relabeled with the predicted pseudo-label $\bar{\y}^{s_i}(h,w)$, \ie
\begin{equation}
\begin{split}
    \hat{\y}^{s_m}(h,w) = 
    q^{\prime}, &\text{ if } \f^{s_m}(h,w,q)>\delta \\ &\text{ and } \bar{\y}^{s_m}(h,w) = q^{\prime}  \text{ and } \y^{s_m}(h,w) = n^{\prime} .\\
\end{split}
\end{equation}

\section{Experiments}
\label{sec:experiments}
We evaluate the effectiveness of our method mDALU under different settings. 
We build benchmarks for image classification, 2D semantic image segmentation, and 2D-3D cross-modal semantic segmentation. 

\subsection{Image Classification}
\textbf{Setup.} In the classification benchmark, 
we adopt the digits classification images from three different datasets, MNIST \cite{lecun2010mnist}, Synthetic Digits \cite{advesarial:alignment:15}, and SVHN \cite{netzer2011reading}, coined ``MT", ``SYN" and ``SVHN", resp. Each time, one of them is taken as the target domain, the other two as source domains. There are 10 classes, from '0' to '9', in the target domain. In our main setting, we adopt the most difficult setup to evaluate different methods, where the label spaces of different source domains are non-overlapping. Only half the classes are labeled in each of the source domains. 
The partially-overlapping situation is also explored.
For fair comparison, we adopt the same network architecture used in~\cite{moment:matching:DA:iccv19} for all methods. The classification performance is evaluated on all 10 classes in the target domain.

\textbf{Comparison with SOTA}. Table \ref{tab:trans_class2} compares our method with other SOTA methods which include 1) unsupervised domain adaptation method DANN \cite{advesarial:alignment:15}, 2) category-shift unsupervised domain adaptation method DCTN \cite{xu2018deep}, 3) multi-source unsupervised domain adaptation method M$^3$SDA \cite{moment:matching:DA:iccv19}, and 4) label unification method AENT \cite{Zhao_UniDet_ECCV20}. It can be seen that without the pseudo label (PL) generation part, other domain adaptation based methods, DANN, DCTN, and M$^3$SDA show the negative transfer effect, or perform similarly to the baseline trained with source data only. This is because each source domain can only provide guidance for a partial label space, and the adaptation in the partial label space guides the prediction on the target domain to the biased label space when training with data from different source domains. This renders the prediction on the target domain contradictory and the model hard to adapt to the complete label space. In contrast, the label-unification based method AENT obtained a performance gain of $4.25\%$, from $60.65\%$ to $64.90\%$, compared with the  source-only baseline. 
This is because it uses an ambiguity cross entropy loss, to avoid the prediction of the source domain data being restricted in a partial label space.

In our first partially-supervised adaptation stage, 
the performance is further improved to $75.13\%$, which proves the effectiveness of our DAT, UM and A$^3$ module for preventing the negative transfer effect. After the second fully-supervised adaptation stage, by adding the PSF module, our model strongly outperforms DCTN \cite{xu2018deep} and AENT \cite{Zhao_UniDet_ECCV20}, both with pseudo-label training, by $15.28\%$ and $9.65\%$, resp. This proves the effectiveness of our entire method for domain adaptation, label space completion and supervision fusion. 
The ablation results in Table~\ref{tab:trans_class_ablation} show that each part of our model contributes to its performance.

\begin{table}
    \centering
    \resizebox{\linewidth}{19.5mm}
    {
    \begin{tabular}{c|cccc}
    \hline
    
    \hline
        Method&MT&SYN&SVHN&Avg\\
        \hline
        Source & 76.76 $\pm$ 0.63 & 61.77 $\pm$ 1.05 & 43.42$\pm$1.89&60.65$\pm$1.19\\
        DANN\cite{advesarial:alignment:15} & 77.30 $\pm$ 2.57 & 60.31 $\pm$ 0.99 & 41.65$\pm$2.34&59.75$\pm$1.97\\
        DANN $^\ast$ & 71.29 $\pm$ 0.48 & 55.94 $\pm$ 0.51 & 35.60 $\pm$ 1.63 & 54.28 $\pm$ 0.87\\
        DCTN \cite{xu2018deep}& 68.10$\pm$0.2 & 62.72$\pm$0.30 & 48.11$\pm$0.57&59.64$\pm$0.36\\
        DCTN $^\ast$ & 72.01 $\pm$ 1.22 & 63.33 $\pm$ 0.20 & 49.34 $\pm$ 1.28 & 61.59 $\pm$ 0.90\\
        M$^3$SDA \cite{moment:matching:DA:iccv19} & 76.56$\pm$0.71 & 61.25$\pm$2.33 & 43.13$\pm$3.55&60.31$\pm$2.20\\
        M$^3$SDA $^\ast$ & 72.50 $\pm$ 2.64 & 55.92 $\pm$ 1.04& 36.24 $\pm$ 1.70 & 54.89 $\pm$ 1.79\\
        AENT\cite{Zhao_UniDet_ECCV20} & 73.24$\pm$1.76 & 68.66$\pm$1.32 & 52.80 $\pm$ 0.92 & 64.90 $\pm$ 1.33\\
        Ours w/o PSF & \textbf{81.23}$\pm$\textbf{0.92} &\textbf{78.97}$\pm$\textbf{0.45}  & \textbf{65.20}$\pm$\textbf{0.58} & \textbf{75.13}$\pm$\textbf{0.65}\\
        \hline
        \hline
        DCTN w/ PL \cite{xu2018deep} & 73.40$\pm$0.85 & 65.63$\pm$0.43 & 52.12$\pm$0.07 & 63.72 $\pm$ 0.45\\
        AENT\cite{Zhao_UniDet_ECCV20} w/ PL & 78.56$\pm$1.23 & 70.25 $\pm$ 0.39 & 59.24 $\pm$ 1.01 & 69.35 $\pm$ 0.88\\
        Ours & \textbf{86.18}$\pm$\textbf{0.45} & \textbf{81.91}$\pm$\textbf{0.33} & \textbf{68.92}$\pm$\textbf{0.81} & \textbf{79.00} $\pm$ \textbf{0.53}\\
        \hline
        
        \hline
    \end{tabular}
    }
    \caption{Quantitative comparison of image classification. ``MT'', ``SYN'', and ``SVHN'' represent the target domain.
    ``PL'' represents to add the pseudo-label training module, which is specifically adjusted according to their own paper's design. $^\ast$ represents to remove the unlabeled samples in the training data. We implement AENT for classification by utilizing the ambiguity cross entropy loss proposed in \cite{Zhao_UniDet_ECCV20}.}
    \label{tab:trans_class2}
    \vspace{-15pt}
\end{table}

\begin{SCfigure}
    \centering
  \includegraphics[width=0.55\linewidth]{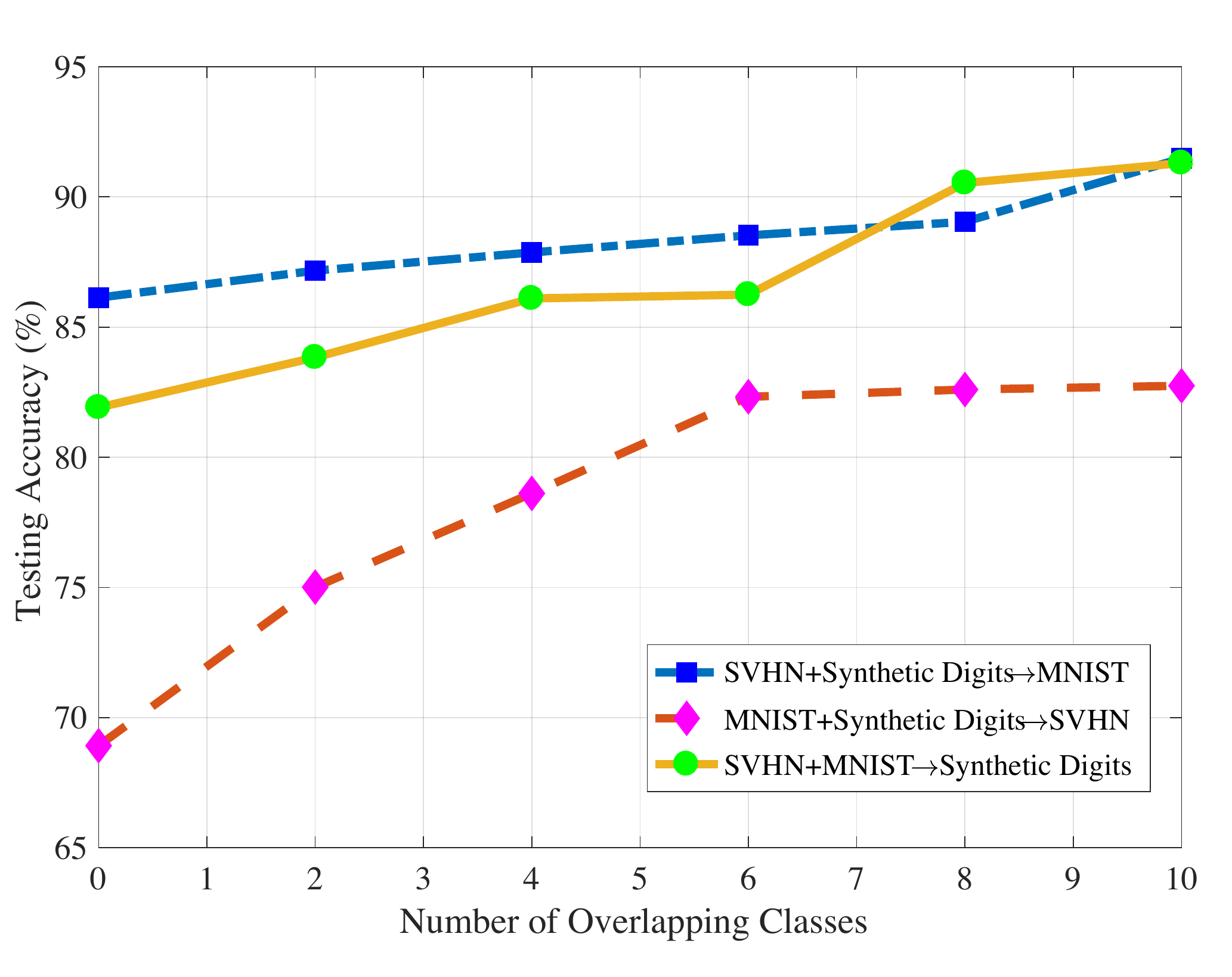}
    \caption{Accuracy in the target domain as a function of the number of overlapping classes between the source domains.}
    \label{fig:overlap}
    \vspace{-10pt}
\end{SCfigure}

\textbf{Partially Overlapping.} 
In Fig. \ref{fig:overlap}, it is shown that the testing accuracy on the target domain increases, as more and more common classes in the source domains are available. 
In Table \ref{tab:trans_class_overlap}, we compare the model performance of our method with other SOTA methods when the source domains are partially overlapping, with 4 common classes. It is shown that our method still strongly outperforms the adaptation-based methods, DANN, DCTN, M$^3$SDA, and the label unification based method, AENT, $82.40\%$ v.s. $69.37\%$, $70.58\%$, $69.56\%$, $76.73\%$. It further verifies the effectiveness of our model in the partial overlap situation.

\begin{table}
\centering
\resizebox{\linewidth}{10.5mm}
    {
\begin{tabular}{cccc|ccccc}
 \hline
 
 \hline
  MBR&UM&A$^3$&PSF&MT&SYN&SVHN&Avg\\
  \hline
  &&&&76.76 $\pm$ 0.63&61.77 $\pm$ 1.05 & 43.42$\pm$1.89&60.65$\pm$1.19\\
  \checkmark&&&&72.21$\pm$1.89 & 62.41$\pm$0.58 & 50.24$\pm$1.23&61.62$\pm$1.23\\
  \checkmark&\checkmark&&&84.74$\pm$0.54 & 76.12$\pm$0.85 & 58.39$\pm$0.57 & 73.08$\pm$ 0.65\\
  \checkmark&\checkmark&\checkmark$^\ast$&&81.38$\pm$0.79 &78.20$\pm$1.3  & 65.12$\pm$0.64 & 74.90 $\pm$ 0.91\\
  \checkmark&\checkmark&\checkmark&&81.23$\pm$0.92 &78.97$\pm$0.45  & 65.20$\pm$0.58 & 75.13 $\pm$ 0.65\\
   \checkmark&\checkmark&\checkmark&\checkmark&\textbf{86.18}$\pm$\textbf{0.45} & \textbf{81.91}$\pm$\textbf{0.33} & \textbf{68.92}$\pm$\textbf{0.81} & \textbf{79.00} $\pm$ \textbf{0.53}\\
  \hline
  
  \hline
 \end{tabular}
 }
\caption{Ablation study under the image classification setting. MBR: multi-branch network, \ie adopts different networks $G_i$ for different source domains. $^\ast$ indicates there is no adversarial part in the A$^3$ module, \ie only the DAT module. The best results are denoted in bold.}
\label{tab:trans_class_ablation}
\vspace{-10pt}
\end{table}

\begin{table}
    \centering
    \resizebox{\linewidth}{11mm}
    {
    \begin{tabular}{c|cccc}
    \hline
    
    \hline
        Method&MT&SYN&SVHN&Avg\\
        \hline
        Source & 82.10$\pm$1.50& 73.37$\pm$ 0.67& 57.50$\pm$1.93 & 70.99 $\pm$ 1.37\\
        DANN\cite{advesarial:alignment:15} & 80.13$\pm$1.60& 72.97$\pm$0.49&55.00$\pm$0.73 & 69.37 $\pm$ 0.94\\
        DCTN\cite{xu2018deep} & 78.56$\pm$0.47& 72.33 $\pm$ 0.04&60.86$\pm$0.21 & 70.58 $\pm$ 0.24\\
        M$^3$SDA\cite{moment:matching:DA:iccv19} & 81.52 $\pm$ 1.55& 72.91 $\pm$ 0.68&54.26$\pm$0.66 & 69.56 $\pm$ 0.96\\
        AENT\cite{Zhao_UniDet_ECCV20} &79.12 $\pm$ 1.07 & 81.99 $\pm$ 0.87 & 69.07 $\pm$ 1.93 & 76.73 $\pm$ 1.29\\
        Ours w/o PSF& \textbf{85.39} $\pm$ \textbf{1.32}& \textbf{85.33}$\pm$ \textbf{1.21}& \textbf{76.48}$\pm$\textbf{1.31} & \textbf{82.40} $\pm$ \textbf{1.28}\\
        \hline
        
        \hline
    \end{tabular}
    }
    \caption{
    Quantitative comparison of image classification, under the partial overlap setting with 4 common classes.
    }
    \label{tab:trans_class_overlap}
    \vspace{-12pt}
\end{table}

\subsection{2D Semantic Image Segmentation}
\textbf{Setup.} In the single mode semantic segmentation setting, 
we adopt the synthetic-to-real image semantic segmentation setup. The synthetic image datasets GTA5 \cite{Richter_2016_ECCV:gta} and the SYNTHIA \cite{RosCVPR16:synthia} are taken as the source domains, while the real image dataset Cityscapes \cite{Cordts2016Cityscapes} is used as the target domain. Information of 19 classes needs to be transferred to the Cityscapes dataset. In our main setting, the label spaces of SYNTHIA and GTA5 are non-overlapping. In the SYNTHIA dataset, the label of 7 classes are available, incl. road, sidewalk, building, vegetation, sky, person and car. In GTA5, the labels of 12 classes are available, being wall, fence, pole, light, sign, terrain, rider, truck, bus, train, motorcycle and bicycle. Furthermore, we also explore the performance of our model when the images of the two source domains are fully labeled. Moreover, we verify the effectiveness of our model when the taxonomies of different source domains are inconsistent. In those inconsistency experiments, for GTA5, the labels wall, fence, pole, light, sign, terrain, truck, bus, train, person (incl. person and rider), cycle (incl. bicycle and motorcycle) are available. In SYNTHIA, the labels road, sidewalk, building, vegetation, sky, person, rider, car, public facilities (incl. wall, fence, pole), motorcycle and bicycle are available. In order to further evaluate the performance of all methods when combined with the pixel-level domain adaptation methods \cite{CycleGAN2017, Hoffman_cycada2017}, we conduct experiments in two settings; 1) source domain images are not translated with CycleGAN \cite{CycleGAN2017}, named as ``NT''; 2) source domain images are translated with CycleGAN, named as ``T''. Also, in order to verify model performance combined with output-level adaptation method \cite{Tsai_adaptseg_2018}, we conduct additional experiments which include ``ADV'' in the fully-supervised adaptation stage. ``ADV'' generates the complete source domain label as in PSF, and then trains the semantic segmentation model via adversarial adaptation between pseudo-complete source domain and unlabeled target domain in the output-level space. For fair comparison, all the methods use the DeepLabv2-ResNet101 \cite{chen2017deeplab,he2016deep:resnet} semantic segmentation network.

\begin{table}
\begin{subtable}{0.45\linewidth}
  \centering
  \resizebox{\linewidth}{14mm}{
  \begin{tabular}{c|cc}
    \hline
    
    \hline
        Method&NT&T\\
        \hline
        Source & 17.7 &  24.0 \\
        AdaptSegNet\cite{Tsai_adaptseg_2018} & 7.7 & 30.8\\
        MinEnt\cite{vu2018advent} & 27.1 & 30.1\\
        Advent\cite{vu2018advent} & 11.8 & 30.3\\
        Ours w/o PSF& \textbf{36.3} & \textbf{38.1}\\
        \hline
        \hline
        Ours (ADV) & 40.1 & 41.5\\
        Ours (PSF) & 37.3 & 42.4\\
        Ours (ADV+PSF) & \textbf{40.6} & \textbf{42.8} \\
        \hline
        
        \hline
    \end{tabular}
    }
    \caption{}
    \label{tab:trans_seg}
  \end{subtable}
  \begin{subtable}{0.49\linewidth}
  \centering
  \resizebox{\linewidth}{14mm}{
  \begin{tabular}{ccccc|cc}
 \hline
 
 \hline
  MBR&UM&A$^3$&PSF&ADV&NT&T\\
  \hline
  &&&&&17.7&24.0\\
  \checkmark&&&&&20.9 & 21.4\\
  \checkmark&\checkmark&&&&27.6 & 36.8\\
  \checkmark&\checkmark&\checkmark$^\ast$&&&29.1  & 37.0\\
  \checkmark&\checkmark&\checkmark&&&36.3 & 38.1\\
  \checkmark&\checkmark&&&\checkmark& 35.4 & 40.9\\
  \checkmark&\checkmark&&\checkmark&& 31.4 & 41.5\\
  \checkmark&\checkmark&\checkmark&&\checkmark& 40.1 & 41.5\\
   \checkmark&\checkmark&\checkmark&\checkmark&&37.3 & 42.4\\
   \checkmark&\checkmark&\checkmark&\checkmark&\checkmark&\textbf{40.6} & \textbf{42.8} \\
  \hline
  
  \hline
 \end{tabular}
    }
    \caption{}
\label{tab:seg_ablation}
  \end{subtable}
  \vspace*{-10pt}
  \caption{(a) Quantitative comparison of single mode semantic segmentation, SYNTHIA+GTA5$\rightarrow$ Cityscapes. 
  The mIoU results are reported for 19 classes. (b) Ablation study for single mode segmentation. $^\ast$ indicates there is no adversarial part in the A$^{3}$ module, \ie only the DAT module.
  ``ADV+PSF'' means to combine ``ADV'' and ``PSF'' by completing the label space and generating pseudo-labels in the source and target domains, then adversarial alignment in the output space is adopted during the second stage training.}
  \vspace{-15pt}
\end{table}

\begin{figure}
    \centering
    \includegraphics[trim=0 1.8cm 0 0,clip,width=\linewidth]{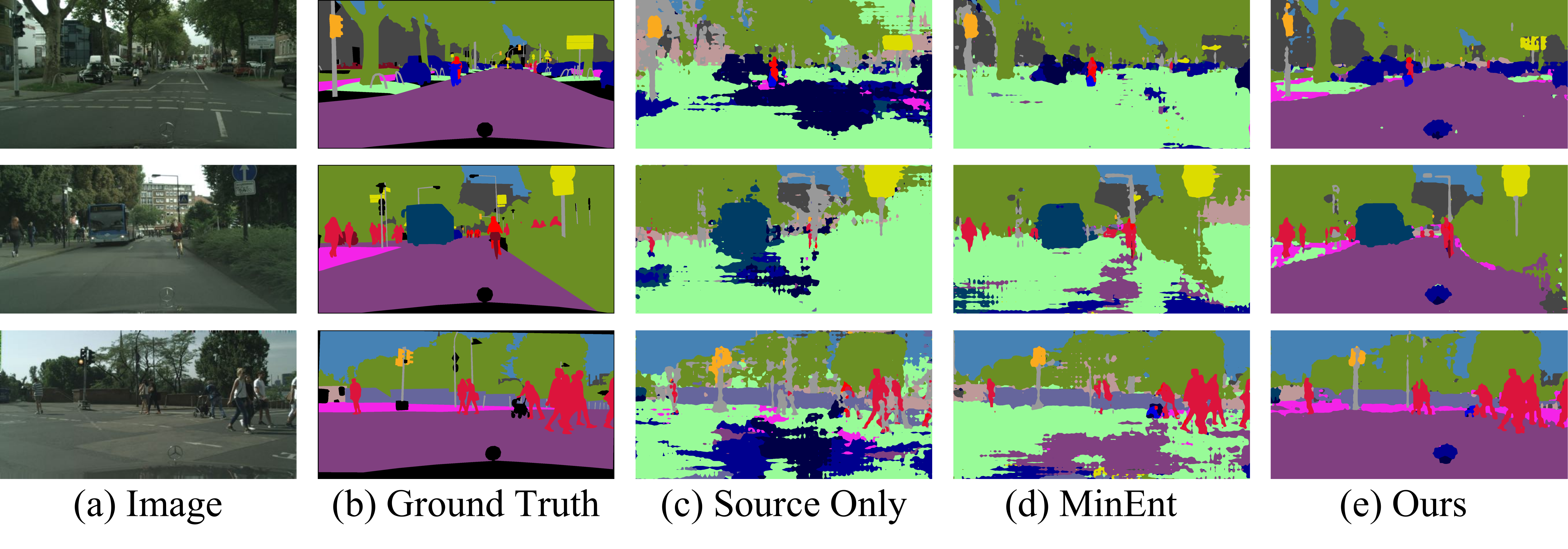}
    \vspace*{-18pt}
    \caption{Qualitative results of 2D semantic segmentation.}
    \label{fig:segvis}
    \vspace{-10pt}
\end{figure}

\textbf{Comparison with SOTA.} In Table \ref{tab:trans_seg}, we show a quantitative comparison for semantic segmentation between our method and other SOTA methods. It is shown that our method without adding PSF strongly outperforms the adaptation-based AdaptSegNet\cite{Tsai_adaptseg_2018}, the self-supervision-based MinEnt\cite{vu2018advent}, and the method combining adaptation and self-supervision Advent \cite{vu2018advent}. Our method achieves  $36.3\%$ and $38.1\%$ in the "NT" and "T" settings, resp. Similar to the image classification results, without using the translated source images, the adaptation-based methods suffer from negative transfer and the performance is lower than the source-only baseline. By using the translated source images in ``T'', different source domain images are all Citysapes-like images. The different source domains can be seen as a larger unified source domain, which can provide guidance for the complete label space to some extent. So all adaptation-based or self-supervision based methods perform much better in the ``T'' situation, compared with the non-adapted baseline. Yet, even in the ``T'' situation, our method still provides an advantage by further completing the label space, through our partially supervised adaptation. This proves the effectiveness of our method in preventing negative transfer and in completing the label space. By further adding the second ``fully-supervised'' adaptation stage, the model achieves a new SOTA performance in both  the ``T'' and the ``NT'' settings. An ablation study, see Table~\ref{tab:seg_ablation}, confirms all parts of our method add to its performance, and the output space alignment ``ADV'' is helpful as well. Fig.~\ref{fig:segvis} shows qualitative results on Cityscapes.

\textbf{Fully labeled.} 
In the fully labeled setting, \ie the source domain images are labeled with all considered classes - 16 classes in SYNTHIA and 19 classes in GTA5 - Table \ref{tab:seg_full_overlap} shows that our model still outperforms other unsupervised domain adaptive semantic segmentation methods, $43.1\%$ vs. $40.8\%$, $42.2\%$, and $42.9\%$. Our model also outperforms the SOTA method for multi-source domain adaptive semantic segmentation MADAN~\cite{zhao2019multi}, $41.9\%$ vs. $41.4\%$.

\begin{table}[]
    \centering
    \resizebox{0.6\linewidth}{15mm}{
     \begin{tabular}{c|c|c|c}
     \hline
     
    \hline
        Method& Base & mIoU$^*$ & mIoU\\
        \hline
        Source & \multirow{5}{*}{\rotatebox{90}{ResNet-101}}& 42.8 & 39.1 \\
        AdaptSegNet\cite{Tsai_adaptseg_2018}& & 45.2 &40.8\\
        Minentropy\cite{vu2018advent}& & 46.4 &42.2\\
        Advent\cite{vu2018advent} && 46.7 &42.9\\
        Ours w/o PSF & & \textbf{46.8} & \textbf{43.1}\\
        \hline\hline
        Source\cite{zhao2019multi} & \multirow{3}{*}{\rotatebox{90}{VGG-16}}& 37.3 & --\\
        MADAN\cite{zhao2019multi}& & 41.4 & --\\
        Ours w/o PSF & &\textbf{41.9} & \textbf{38.0}\\
        \hline
        
        \hline
    \end{tabular}
    }
    \caption{Single mode segmentation results, under fully-labeled setting and ``T". mIoU$^*$ is the mean IoU of 16 classes in SYNTHIA, while mIoU is that of all 19 classes.}
    \vspace{-10pt}
    \label{tab:seg_full_overlap}
\end{table}

\textbf{Inconsistent Taxonomies.} 
Table \ref{tab:seg_incon_taxnon} shows that our method is advantageous when taxonomies are inconsistent, $40.0\%$ vs. $28.1\%$, $31.9\%$, $32.2\%$. In the partially supervised adaptation stage, as in Sec.~\ref{sec:inconsistent_taxonomy}, by adding higher weights to ``person", ``rider", ``motorcycle" and ``bicycle" for SYNTHIA and ``wall", ``fence" and ``pole" for GTA5, our method can achieve a higher performance than inference without weighting, $37.2\%$ vs. $35.3\%$. After the fully supervised adaptation stage, the performance can be further improved to $40.0\%$. The detailed performance for inconsistent taxonomies classes in Table \ref{tab:seg_incon_taxnon} underlines the effectiveness of our method for the inconsistent taxonomies.

\begin{table}[]
    \centering
    \resizebox{\linewidth}{17.8mm}{
     \begin{tabular}{c|ccccccc|c}
     \hline
     
    \hline
        Method&\rotatebox{90}{wall}&\rotatebox{90}{fence}&\rotatebox{90}{pole}&\rotatebox{90}{person}&\rotatebox{90}{rider}&\rotatebox{90}{motorcycle}&\rotatebox{90}{bicycle}&mIoU\\
        \hline
        Source & 2.6 & 12.0 & 12.3 & 40.6 & 0.5 & 0.1 & 28.6 &19.8 \\
        AdaptSegNet\cite{Tsai_adaptseg_2018} & 7.1 & 2.6 & 4.0 & 33.2 & 6.9 & 1.8 & 37.6 &28.1\\
        Minentropy\cite{vu2018advent} & 6.7 & \textbf{18.1} & 23.0 & 28.8 & 6.6 & 1.0 & 42.3 &31.9\\
        Advent\cite{vu2018advent} &6.2 & 11.5 & 11.4 & 32.8 & 12.2 & 0.9 & 41.2 &32.2\\
        Ours w/o PSF & 12.3 & 15.2 & 21.2 & 48.4 & 3.3 & 1.3 & 42.4 &35.3 \\
        Ours w/o PSF $^*$ & \textbf{14.1} & 15.3 & \textbf{30.6} & 48.1 & 17.9 & 13.0 & 42.1 & 37.2\\
        Ours (PSF) & 13.3 & 17.9 & \textbf{30.6} & \textbf{53.7} & \textbf{18.2} & \textbf{19.8} & \textbf{43.2} &\textbf{40.0}\\
        \hline
        
        \hline
    \end{tabular}
    }
    \vspace*{-5pt}
    \caption{Quantitative comparison of single mode segmentation, with inconsistent taxonomies, in the ``T" setting. $^*$During inference, an additional weights map is adopted in case of inconsistent taxonomies as in Sec. \ref{sec:inconsistent_taxonomy}. The detailed performance on inconsistent taxonomies classes is also shown. The mIoU is reported for 19 classes.}
    \label{tab:seg_incon_taxnon}
    \vspace{-10pt}
\end{table}

\subsection{Cross-Modal Semantic Segmentation}
\textbf{Setup.} In the cross-modal semantic segmentation setting, 
the 2D RGB images from Cityscapes  \cite{Cordts2016Cityscapes}, and the 3D LiDAR point clouds from Nuscenes \cite{nuscenes2019} are treated as two different source domains, while the paired but unlabeled 2D RGB images and 3D point clouds from A2D2 \cite{geyer2020a2d2} are used as the target domain. There are 10 classes in total that need to be transferred to the target domain. In Cityscapes, the label for 6 classes are given, covering road, sidewalk, building, pole, sign and nature. In Nuscenes the labels for 4 classes are given, incl. person, car, truck and bike. The 2D RGB images and 3D point clouds in the target domain are registered via a projection matrix between the 2D pixel and 3D points. Following \cite{Jaritz_2020_CVPR}, we adopt U-Net-ResNet34 \cite{ronneberger2015u:unet, he2016deep:resnet} as the 2D semantic segmentation network, and SparseConvNet \cite{SparseConvNet} for 3D semantic segmentation. Due to the challenge of aligning features for the 3D point clouds, the A$^3$ module is not included in the cross-modal setting.

\begin{table}
    \centering
    \resizebox{0.8\linewidth}{16mm}
    {
    \begin{tabular}{c|cc|c}
    \hline
    
    \hline
        Cityscapes + Nuscenes $\rightarrow$ A2D2 &2D&3D&Fuse\\
        \hline
        Source & 37.5 & 2.0 & 42.5 \\
        xMUDA\cite{Jaritz_2020_CVPR} & 16.3 & 1.7 & 9.1 \\
        ES + MinEnt\cite{vu2018advent} & 22.3 & 1.5 & 20.8 \\
        ES + KL\cite{Jaritz_2020_CVPR} & 21.7 & 1.5 & 19.7 \\
        xMUDA + AKL & 27.5& 2.3 & 21.1 \\ 
        xMUDA + AKL + COMP & 32.1 & 2.9 &37.7 \\
        \hline
        \hline
        Ours w/o PSF & 38.1 & 2.4 & 49.9\\
        Ours & \textbf{54.9} & \textbf{37.1} & \textbf{55.7}\\
        \hline
        
        \hline
    \end{tabular}
    }
    \vspace*{-5pt}
    \caption{Quantitative comparison of cross modal segmentation, Nuscenes+Cityscapes$\rightarrow$ A2D2. "Fuse" represents the average fusion of the prediction probability from 2D models and 3D models; the final class prediction is the maximum of the fused probability. ``ES" means 2D and 3D average fusion ensemble. ``KL" means KL-divergence alignment. ``AKL" means adaptive KL-divergence alignment. ``COMP" means complementary condition constraint for the point. The mIoU is reported over 10 classes on A2D2.}
    \label{tab:seg_cross}
    \vspace{-10pt}
\end{table}

\begin{figure}
    \centering
    \includegraphics[trim=0 0.4cm 0 0, clip, width=\linewidth]{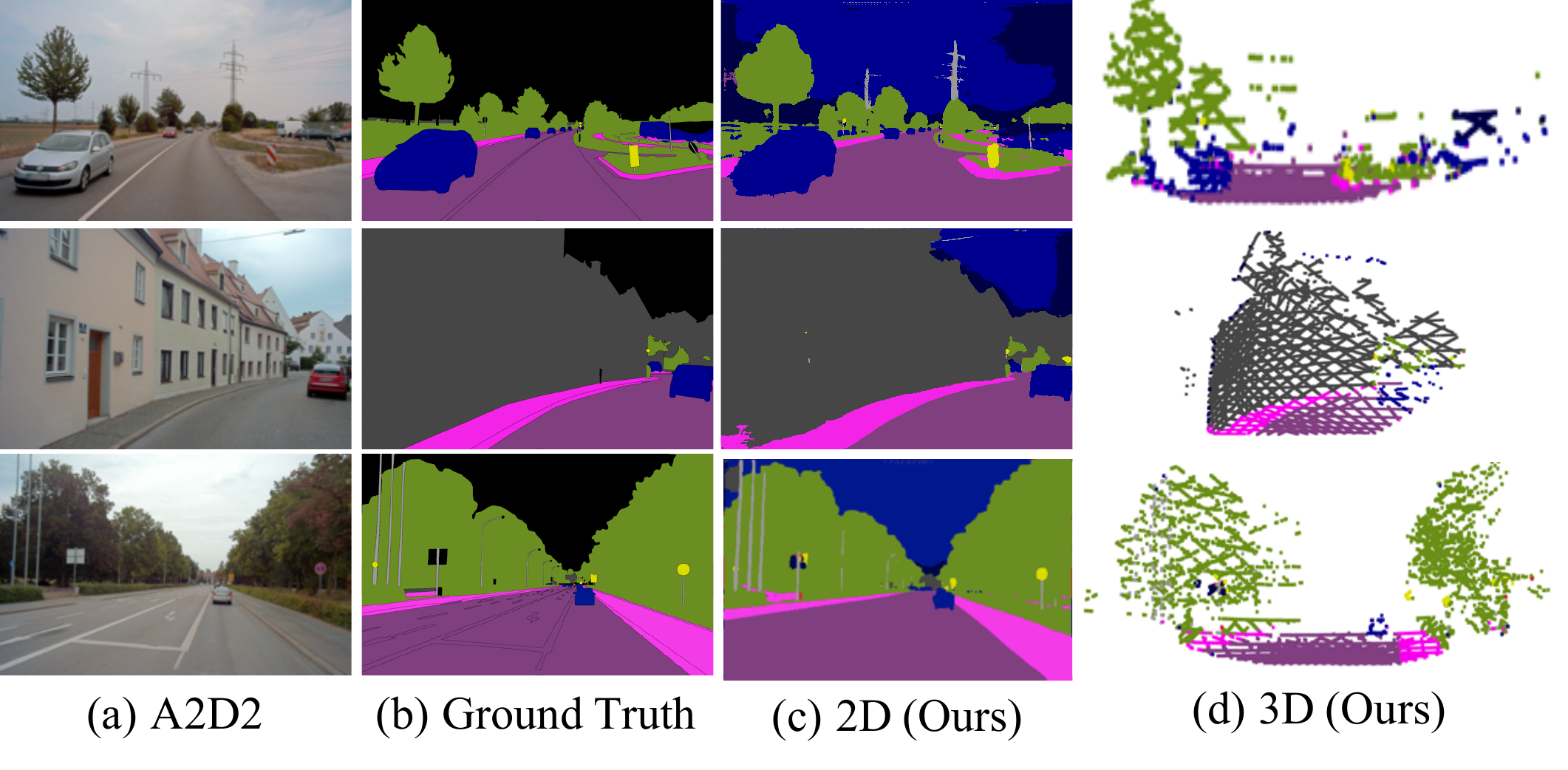}
    \vspace*{-18pt}
    \caption{Qualitative results of the cross-modal setting.}
    \label{fig:cross_visual}
    \vspace{-15pt}
\end{figure}

\textbf{Comparison with the SOTA.} 
As shown in Table \ref{tab:seg_cross}, similar to the image classification and the single mode semantic segmentation results, the SOTA cross-modal unsupervised adaptation method xMUDA~\cite{Jaritz_2020_CVPR} shows an obvious negative transfer effect, resulting in a performance drop for the 2D model, 3D model and the fused one. Furthermore, we designed reasonable baseline methods for comparison: 1) ES + MinEnt: the prediction from 2D and 3D networks are averaged in the target domain through the 2D and 3D point correspondence during training, and the fused prediction probability is optimized using the minimum entropy loss \cite{vu2018advent}. 2) ES + KL: the KL-divergence \cite{Jaritz_2020_CVPR} is utilized to align between the 2D/3D prediction and the fused predictions for the corresponding points in the target domain, resp. 3) xMUDA + AKL: the KL-divergence alignment between 2D and 3D in the target domain is weighted adaptively, to reduce the wrong guidance from the unlabeled parts. 4) xMUDA + AKL + COMP: following baseline 3), another constraint, that the weights related to 2D and 3D need to be complementary, is added. It is shown that our method prevents negative transfer without the PSF component, outperforming the non-adapted baseline. Then by adding the PSF module, the 2D and 3D single-model performance is strongly improved, achieving $54.9\%$ and $37.1\%$, resp. In Fig.~\ref{fig:cross_visual}, we show qualitative results in the target domain. The good performance proves the effectiveness of our method for the mDALU with partial modalities. This opens up the avenue to combine datasets collected with different sensors and offers the possibility of cheaply evaluating new combinations of sensors without annotating their data.

\vspace{-5pt}\section{Conclusion}
In this paper, we proposed the multi-source domain adaptation and label unification with partial datasets problem, called mDALU. Then we proposed a novel multi-stage approach for mDALU, including partially and fully supervised adaptation stages. Our approach is demonstrated through extensive experiments on different benchmarks.

\textbf{Acknowledgements.} This research has received funding from the EU Horizon 2020 research and innovation programme under grant agreement No. 820434. Dengxin Dai is the corresponding author.

{\small
\bibliographystyle{ieee_fullname}
\bibliography{egbib}
}
\newpage
\clearpage

\setcounter{equation}{0}
\setcounter{figure}{0}
\setcounter{table}{0}
\setcounter{page}{1}
\setcounter{section}{0}

\renewcommand{\thesection}{S\arabic{section}}  
\renewcommand{\thefigure}{S\arabic{figure}}
\renewcommand{\thetable}{S\arabic{table}}
\maketitle
\ificcvfinal\thispagestyle{empty}\fi
\section*{Supplementary}

In this supplementary, we provide additional information for,
\begin{itemize}
\item[\textbf{S1}] detailed framework structure and implementation of our approach,
\item[\textbf{S2}] more detailed information about the datasets involved in experiments,
\item[\textbf{S3}] experimental results when having more than two source domains,
\item[\textbf{S4}] more experimental results and additional visualization results  for semantic segmentation.

\end{itemize}

\section{Framework Structure and Implementation}
\begin{figure*}
    \centering
    \includegraphics[width=\linewidth]{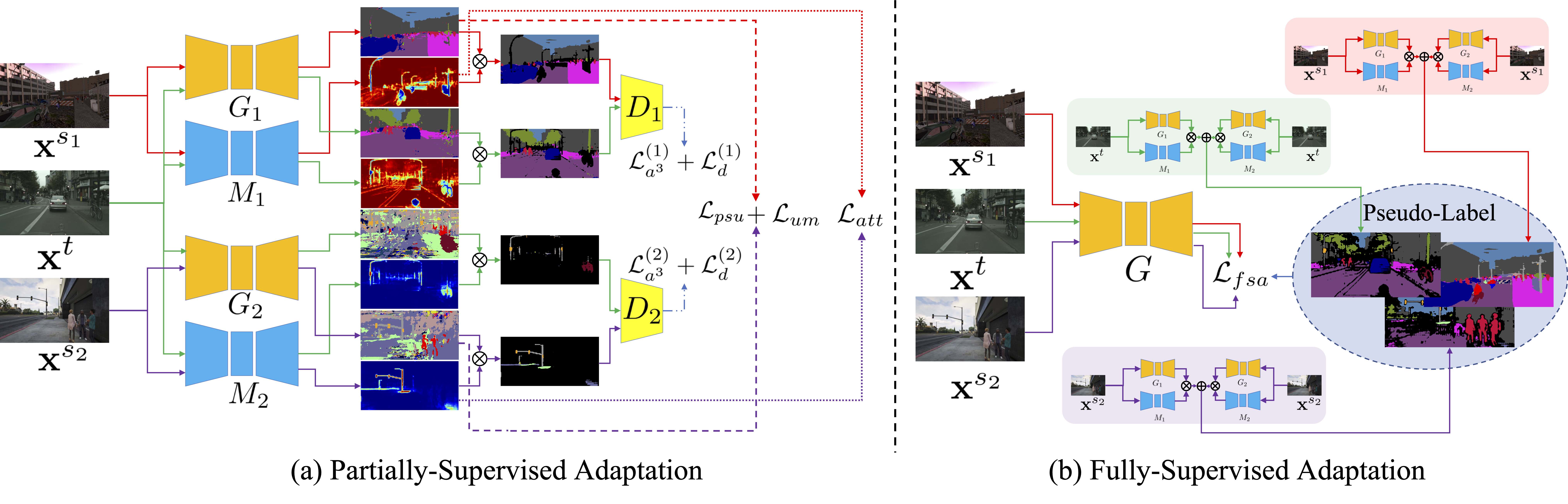}
    \caption{Overview of our approach to mDALU problem. Our approach is composed of two stages: (a) partially-supervised adaptation stage, and (b) fully-supervised adaptation stage. In the partially-supervised adaptation stage, there are three modules involved, the domain attention (DAT) module, the uncertainty maximization (UM) module, and the attention-guided adversarial alignment (A$^3$) module. Besides the supervised semantic segmentation loss $\cL_{psu}$ on the source domain, the DAT module is trained in the supervised way with $\cL_{att}$, the UM module is trained in the supervised way with $\cL_{um}$ and the A$^3$ module is trained in the adversarial way with $\cL_{a^3}+\cL_{d}$. In the fully-supervised adaptation stage, in order to complete the label space, the pseudo-label, for all the samples $\x^{s_1}$, $\x^{s_2}$, $\x^{t}$ from all related domains, is generated by fusing the probability map weighted by attention map from different branches, $G_1, M_1$ and $G_2, M_2$. Then the semantic segmentation network $G$ is trained in the complete and unified label space with the generated pseudo-label and the supervised loss $\cL_{fsa}$. In the implementation, $G_1, G_2, M_1, M_2$ share the same encoder and adopt different label predictors.}
    \label{fig:overview_approach}
\end{figure*}

In Sec. \textcolor{red}{3} and Fig. \textcolor{red}{2} of the main paper, we introduce our approach to mDALU problem, and here we provide more detailed structure and implementation of our approach. The overview of our approach is shown in Fig. \ref{fig:overview_approach}. In the image classification experiment, the hyperparameter $\lambda$ in Eq.~(\textcolor{red}{10}) of the main paper is set as 1.0, and $\delta$ in Eq.~(\textcolor{red}{12}) and Eq.~(\textcolor{red}{13}) of the main paper is set as 0.5. The images are resized to $32\times32$. We use the the Adam optimizer~\cite{kingma2014adam} with $\beta_1=0.9, \beta_2=0.999$ and the weight decay as $5\times 10^{-4}$. The learning rate is set as $2\times 10^{-4}$. We adopt the same network architecture as that of the digits classification experiments in \cite{moment:matching:DA:iccv19}. In the 2D semantic image segmentation experiments, the hyperparameter $\lambda$ in Eq. (\textcolor{red}{10}) of the main paper is set as 0.001, and $\delta$ in Eq. (\textcolor{red}{12}) and Eq. (\textcolor{red}{13}) of the main paper is set as 0.2, 0.5 and 0.4 for SYNTHIA, GTA5 and Cityscapes dataset, respectively. The images are resized to $1024\times 512$. We use the SGD optimizer for training the semantic segmentation network, whose momentum is 0.9, weight decay is $5\times 10^{-4}$ and learning rate is $2.5\times 10^{-4}$ with polynomial decay of power 0.9. Meanwhile, the Adam optimizer is used for training the discriminator network, whose momentum is $\beta_1=0.9, \beta_2=0.99$, weight decay is $5\times 10^{-4}$ and learning rate is $1\times 10^{-4}$ with polynomial decay of power 0.9. We adopt the same semantic segmentation and discriminator network architecture as that of~\cite{Tsai_adaptseg_2018}. In the cross-modal semantic segmentation experiments, we follow the exactly same data augmentation and preprocess procedure as that of \cite{Jaritz_2020_CVPR}. The hyperparameter $\delta$ in Eq. (\textcolor{red}{12}) and Eq. (\textcolor{red}{13}) of the main paper is set as 0.2. We use the Adam optimizer for training the 2D and 3D semantic segmentation network, with $\beta_1=0.9, \beta_2=0.999$. The learning rate is set as $1\times 10^{-3}$.

\section{Datasets Overview of mDALU Benchmark}
In Sec. \textcolor{red}{4} of the main paper, we introduce the benchmark setup of the mDALU problem. Here we provide more details about the datasets involved in the benchmark. 
\subsection{Image Classification}
\begin{figure}
    \centering
    \includegraphics{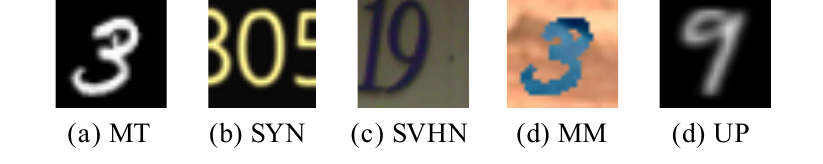}
    \caption{Example images of different datasets in mDALU image classification benchmark.}
    \label{fig:classification_example}
\end{figure}

In the image classification benchmark of the main paper, we adopt three digits datasets, including MNIST \cite{lecun2010mnist}, Synthetic Digits \cite{ganin2015unsupervised}, and SVHN \cite{netzer2011reading} dataset. MNIST is a hand-written numbers image dataset, SVHN is a street view house numbers image dataset and Synthetic Digits is a synthetic numbers image dataset. In the image classification benchmark of the main paper, we adopt these three different style digits images, to introduce larger domain gap between different source domains to effectively evaluate the validity of different methods for mDALU problem. In Sec. \ref{sec:more_source}, we introduce two more datasets, MNIST-M~\cite{ganin2015unsupervised} and USPS~\cite{hull1994database} , to evaluate the effectiveness of our approach when dealing with more than two source domains. MNIST-M is a synthetic numbers image dataset, and USPS is a hand-written numbers image dataset. We follow the setup of splitting the dataset in \cite{moment:matching:DA:iccv19, peng2019federated}. In each of MNIST, MNIST-M, SVHN and Synthetic Digits, 25000 images for training are sampled from the training subset, and 9000 images for testing are sampled from the testing subset. And for the USPS dataset, due to there are only 9298 images in total are available, the whole training set covering 7438 images is used for training, while the whole testing set with 1860 images is adopted for testing. MNIST, MNIST-M, SVHN, Synthetic Digits, USPS are abbreviated as MT, MM, SVHN, SYN, and UP, respectively. The detailed label space of different source domains and the target domain under different experiments setup is listed in Table \ref{tab:label_space_main} and Table \ref{tab:label_space_supp}. The example images of different datasets are shown in Fig. \ref{fig:classification_example}. 

\begin{table*}[!ht]
    \centering
    \resizebox{\textwidth}{!}{
    \begin{tabular}{c|c|ccc|ccc|ccc}
    \hline
    
    \hline
    Experiment & \multicolumn{10}{c}{Label Space} \\
        
        \hline
        \multirow{3}{*}{Non-Overlapping(Table \textcolor{red}{2} in main paper)} & 
        Domain & Source1 & Source2 & Target & Source1 & Source2 & Target & Source1 & Source2 & Target\\
        & Dataset & SVHN & SYN & MT & MT & SVHN & SYN & MNIST & SYN & SVHN\\
         & Class & 0$\sim$4 & 5$\sim$9 & 0$\sim$9 & 0$\sim$4 & 5$\sim$9 & 0$\sim$9 & 0$\sim$4 & 5$\sim$9 & 0$\sim$9\\
        \hline
        
        \hline
        \multirow{3}{*}{Partially-Overlapping(Table \textcolor{red}{4} in main paper)} &
        Domain & Source1 & Source2 & Target & Source1 & Source2 & Target & Source1 & Source2 & Target\\
        & Dataset & SVHN & SYN & MT & MT & SVHN & SYN & MNIST & SYN & SVHN\\
        & Class & 0$\sim$6 & 3$\sim$9 & 0$\sim$9 & 0$\sim$6 & 3$\sim$9 & 0$\sim$9 & 0$\sim$6 & 3$\sim$9 & 0$\sim$9\\
        \hline
        
        \hline
    
    \end{tabular}
    }
    \caption{The label space of different source domains and the target domain in the mDALU image classification benchmark of the main paper.}
    \label{tab:label_space_main}
\end{table*}

\begin{table*}[!ht]
    \centering
    \resizebox{0.6\textwidth}{!}{
    \begin{tabular}{c|ccccc}
    \hline
    
    \hline
    \multicolumn{6}{c}{More Source Domains Experiments (Table \ref{tab:more_source_supp} in supplementary)} \\
    \hline
    
    \hline
        Domain & Source1 & Source2 & Source3 & Source4 & Target \\
        \hline
        Dataset & SVHN & SYN & MM & UP & MT\\
        Class & 0$\sim$2 & 2$\sim$4 & 4$\sim$6 & 7$\sim$9 & 0$\sim$9\\
        Dataset & MT & SYN & MM & UP & SVHN\\
        Class & 0$\sim$2 & 2$\sim$4 & 4$\sim$6 & 7$\sim$9 & 0$\sim$9\\
        Dataset & MT & SVHN & MM & UP & SYN\\
        Class & 0$\sim$2 & 2$\sim$4 & 4$\sim$6 & 7$\sim$9 & 0$\sim$9\\
        Dataset & MT & SVHN & SYN & UP & MM\\
        Class & 0$\sim$2 & 2$\sim$4 & 4$\sim$6 & 7$\sim$9 & 0$\sim$9\\
        Dataset & MT & SVHN & SYN & MM & UP\\
        Class & 0$\sim$2 & 2$\sim$4 & 4$\sim$6 & 7$\sim$9 & 0$\sim$9\\
        \hline
        
        \hline
    \end{tabular}
    }
    \caption{The label space of different source domains and the target domain in the mDALU image classification benchmark of the more source domains experiments in the supplementary.}
    \label{tab:label_space_supp}
\end{table*}

\subsection{2D Semantic Image Segmentation} \label{sec:2d_seg_supp}
In the 2D semantic image segmentation benchmark of the main paper, we adopt the synthetic image datasets, GTA5 \cite{Richter_2016_ECCV:gta} and SYNTHIA \cite{RosCVPR16:synthia} and the real image dataset, Cityscapes \cite{Cordts2016Cityscapes}. We introduce the label space of different datasets in the main paper. Here we provide more additional information about the datasets.

\textbf{Cityscapes.} Cityscapes is a dataset composed of the street scene images collected from different European cities. We use the training set of Cityscapes covering 2993 images, without the label information, as the target domain during the training stage. And we adopt the validation set of Cityscapes, which are composed of 500 images and densely labeled with 19 classes, to evaluate the semantic segmentation performance of the model on the target domain.

\textbf{GTA5.} GTA5 is a synthetic urban scene image dataset, whose images are rendered from the game engine. The scene of the images is based on the city of Los Angeles. In our 2D semantic image segmentation benchmark, we use 24966 densely labeled images in the GTA5 dataset as one of our source domains, whose annotation is compatible with that of Cityscapes.

\textbf{SYNTHIA.} SYNTHIA is a synthetic dataset, containing photo-realistic images rendered from a virtual city. We use the SYNTHIA-RAND-Cityscapes subset, which contains 9400 densely labeled images and the 16 class annotation of which is compatible with that of Cityscapes. In our 2D semantic image segmentation benchmark, the labeled SYNTHIA dataset serves as one of our source domains.

\subsection{Cross-Modal Semantic Segmentation}
In the cross-modal semantic segmentation benchmark of the main paper, three datasets are involved, Cityscapes \cite{Cordts2016Cityscapes}, Nuscenes \cite{nuscenes2019} and A2D2 \cite{geyer2020a2d2}. We introduce the label space of different datasets in the main paper. Here we provide more information on the datasets and the mapping between our label space and the annotated class label in different datasets. 

\textbf{Cityscapes.} Cityscapes \cite{Cordts2016Cityscapes} is a 2D urban scene image dataset, and has been introduced in the Sec. \ref{sec:2d_seg_supp}. In the cross-modal semantic segmentation benchmark, we adopt the training set of Cityscapes, covering 2975 images, as the 2D source domain. Unlike the Sec. \ref{sec:2d_seg_supp} does not use the label information of Cityscapes training images, we use the ground truth label of Cityscapes training images, but the label space of Cityscapes in our experiments only covers 6 classes, road, sidewalk, building, pole, sign and nature. The mapping from the original Cityscapes annotated classes and our label space is listed in Table \ref{tab:cross_class_mapping}. 

\textbf{Nuscenes.} Nuscenes \cite{nuscenes2019} is an autonomous driving dataset covering 1000 driving scenes, which are collected from the Boston and Singapore. Each scene, of 20-second length, is sampled and annotated at 2HZ, resulting in 40K well-annotated keyframes for 3D bounding boxes of the objects. In our cross-modal semantic segmentation benchmark, we adopt the training set of the Nuscenes, including 28130 keyframes 3D LiDAR points, as the 3D source domain. Then as done in \cite{Jaritz_2020_CVPR}, we generate the 3D point-wise semantic labels from the 3D bounding boxes, by assigning the object label to the points inside the bounding box and taking the points outside the bounding box as unlabeled points. The label space of the 3D source domain includes 4 classes, person, car, truck and bike. The mapping between the object label annotation in Nuscenes and our label space is reported in Table \ref{tab:cross_class_mapping}. 

\begin{figure}
    \centering
    \includegraphics[width=\linewidth]{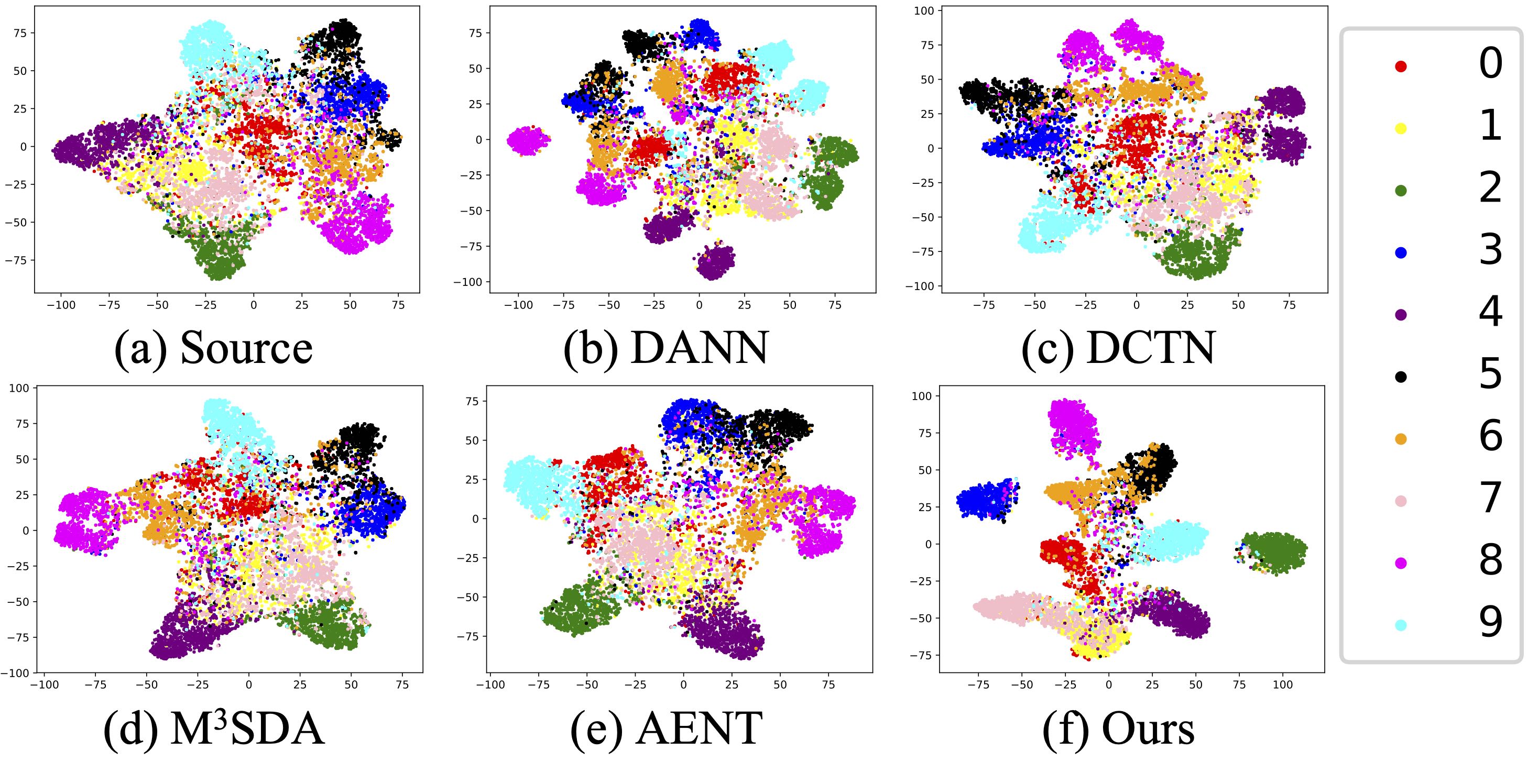}
    \caption{t-SNE Visualization of the feature embedding on the mDALU image classification benchmark, MT, SVHN, MM, UP $\rightarrow$ SYN. We adopt the same t-SNE parameters for all visualization.}
    \label{fig:tsneclassification}
    \vspace{-10pt}
\end{figure}

\textbf{A2D2.} A2D2 \cite{geyer2020a2d2} is an autonomous driving dataset, including simultaneously recorded paired 2D images and 3D LiDAR points. The A2D2 covers 20 scenes, which are corresponding to 28637 frames for training. And the scene 20180807$\_$145028 is used for validation. The 2D images are densely labeled with 38 semantic classes. Following~\cite{Jaritz_2020_CVPR}, the 3D point-wise semantic labels are generated by the reprojection to the 2D images. In our cross-modal semantic segmentation benchmark, the A2D2 serves as the target domain. We use the training set of A2D2 without the label information during training, including the paired 2D images and 3D LiDAR points. And we use the validation set 20180807$\_$145028 with the ground truth label for evaluating the performance. The label space of the target domain for evaluation includes 10 classes, road, sidewalk, building, pole, sign, nature, person, car, truck and bike. The mapping between the label space and the annotated 38 semantic classses in A2D2 is shown in Table \ref{tab:cross_class_mapping}.

\begin{table}[]
    \centering
    \resizebox{\linewidth}{9.2mm}{
     \begin{tabular}{c|ccccccc|c}
     \hline
     
    \hline
        Method&\rotatebox{90}{wall}&\rotatebox{90}{fence}&\rotatebox{90}{pole}&\rotatebox{90}{person}&\rotatebox{90}{rider}&\rotatebox{90}{motorcycle}&\rotatebox{90}{bicycle}&mIoU\\
        \hline
        Ours (PSF w/o relabeling) & 11.5 & 17.8 & \textbf{33.6} & 47.7 & 13.2 & 7.2 & \textbf{43.4} & 38.4 \\
        Ours (PSF w/ relabeling) & \textbf{13.3} & \textbf{17.9} & 30.6 & \textbf{53.7} & \textbf{18.2} & \textbf{19.8} & 43.2 &\textbf{40.0}\\
        \hline
        
        \hline
    \end{tabular}
    }
    \caption{Quantitative comparison of w/ and w/o relabeling inconsistent taxonomies in PSF module. The detailed performance on inconsistent taxonomies classes is also shown. The mIoU is reported for 19 classes. The best results are denoted in bold.}
    \label{tab:seg_icon_relabel}
    \vspace{-10pt}
\end{table}

\begin{table*}[]
    \centering
    \resizebox{\textwidth}{!}{
    \begin{tabular}{c|c|c|c}
    \hline
         label space & A2D2 & Cityscapes & Nuscenes\\
         \hline
         road & \makecell[c]{`rd normal street', `zebra crossing', `solid line', `rd restricted area', `slow drive area',\\ `drivable cobblestone', `dashed line', `painted driv. instr.'} & `road' & --\\
         \hline
         sidewalk & `sidewalk', `curbstone' & `sidewalk' & --\\
         \hline
         building & `buildings' & `building' & --\\
         \hline
         pole & `poles' & `pole' & --\\
         \hline
        sign & `traffic sign 1', `traffic sign 2', `traffic sign 3' & `traffic sign' & --\\
        \hline
        nature & `nature object' & `vegetation', `terrain' & --\\
        \hline
        person & `pedestrian 1', `pedestrian 2', `pedestrian 3' & -- & `pedestrian'\\
        \hline
        car & `car 1', `car 2', `car 3', `car 4', `ego car' & -- & `car'\\
        \hline
        truck & `truck 1', `truck 2', `truck 3' & -- & `truck' \\
        \hline
        bike & \makecell[c]{`bicycle 1', `bicycle 2', `bicycle 3', `bicycle 4', \\ `small vehicles 1', `small vehicles 2', `small vehicles 3'} & -- & `motorcycle', `bicycle'\\
         
    \hline     
    \end{tabular}
    }
    \caption{Class mapping between the label space and the annotated classes in different datasets. }
    \label{tab:cross_class_mapping}
\end{table*}

\section{Experiments with More Source Domains} \label{sec:more_source}
In this section, we evaluate the effectiveness of our approach when dealing with more than two source domains. 
Based on the classification benchmark of the main paper, we here introduce two more datasets, MNIST-M \cite{ganin2015unsupervised} and USPS~\cite{hull1994database}, which are abbreviated as ``MM'' and ``UP'' respectively. Then as done in the main paper, each time, one of ``MT", ``SYN'', ``SVHN'', ``'MM' and ``UP'' is taken as the target domain, while the other four are used as source domains. The label space of different source domains in the experiments is listed in Table \ref{tab:label_space_supp}.

\textbf{Experimental results.} In Table \ref{tab:more_source_supp}, we report the quantitative experimental results of the classification benchmark, after introducing two more datasets, MM and UP. It can be seen that our approach with the ``partially-supervised adaptation" stage highly outperforms the source-only baseline, the adaptation-based methods DANN, DCTN, and M$^3$SDA, and the label-unification based method AENT. It achieves an average accuracy of $80.83\%$ on the target domain. Then by exploiting the ``fully-supervised adaptation" stage, the performance is further improved to $82.88\%$. It proves the effectiveness and the robustness of our approach for addressing the mDALU problem when more than two source domains are given. In Fig. \ref{fig:tsneclassification}, the qualitative comparison of feature embedding, t-SNE visualization~\cite{maaten2008visualizing}, between our approach and other methods is shown. It shows that our approach is able to learn more discriminative features than other methods. It further verifies the good performance of our approach to mDALU problem.

\begin{table*}
    \centering
    \setlength{\tabcolsep}{5pt}
    \begin{tabular}{c|cccccc}
    \hline
        Method&MT&SYN&SVHN&MM&UP&Avg\\
        \hline
        Source & 86.90 $\pm$ 0.40 & 63.80 $\pm$ 0.15 & 51.84 $\pm$ 2.13 & 52.09 $\pm$ 0.69 & 91.83 $\pm$ 0.78 & 69.29 $\pm$ 0.83\\
        DANN\cite{advesarial:alignment:15} & 86.38 $\pm$ 1.44 & 63.76 $\pm$ 0.88 & 51.58 $\pm$ 2.27 & 52.14 $\pm$ 0.61 & 89.98 $\pm$ 1.42 & 68.77 $\pm$ 1.32\\
        DCTN \cite{xu2018deep}& 63.87 $\pm$ 0.10 & 53.33 $\pm$ 1.15 & 43.57 $\pm$ 0.98 & 40.23 $\pm$ 0.48 & 59.78 $\pm$ 1.19 & 52.16 $\pm$ 0.78\\
        M$^3$SDA \cite{moment:matching:DA:iccv19} & 87.26 $\pm$ 1.54 & 63.40 $\pm$ 0.32 & 48.96 $\pm$ 0.92 & 52.28 $\pm$ 1.60 & 90.20 $\pm$ 0.97 & 68.42 $\pm$ 1.07\\
        AENT\cite{Zhao_UniDet_ECCV20} & 79.55 $\pm$ 2.40 & 63.22 $\pm$ 0.41 & 52.58 $\pm$ 2.27 & 48.65 $\pm$ 0.31 & 87.62 $\pm$ 1.36 & 66.32 $\pm$ 1.35\\
        Ours w/o PSF & \textbf{94.90$\pm$0.23} & \textbf{78.37$\pm$0.58} & \textbf{72.18$\pm$0.44} & \textbf{63.01$\pm$0.74} & \textbf{95.70$\pm$0.44} & \textbf{80.83$\pm$0.49}\\
        \hline
        \hline
        Ours & \textbf{96.60$\pm$0.07} & \textbf{80.68$\pm$0.30} & \textbf{73.82$\pm$0.35} & \textbf{66.62$\pm$0.62} & \textbf{96.70 $\pm$ 0.22} & \textbf{82.88$\pm$0.31}\\
        \hline
    \end{tabular}
    \caption{Quantitative comparison between our method and other SOTA methods, under mDALU image classification benchmark with 4 source domains. ``MT'', ``SYN'', ``SVHN'', ``MM'', and ``UP'' represent the target domain. We implement AENT on classification by utilizing the ambiguity cross entropy loss proposed in \cite{Zhao_UniDet_ECCV20}. The best results are denoted in bold.}
    \label{tab:more_source_supp}
\end{table*}

\section{More Experimental Results for Semantic Segmentation}

\textbf{Detailed experimental results for semantic segmentation.}
In Table \textcolor{red}{5a} and Table \textcolor{red}{8} of the main paper, we show the quantitative comparison, through the mIoU, between our approach and other methods, on the 2D and cross-modal semantic segmentation benchmark. Correspondingly, we here provide more detailed experimental results in Table \ref{tab:seg_2d_per_class} and Table \ref{tab:cross_perclass}, covering the per-class IoU results. 

\begin{table*}
    \centering
    \resizebox{\textwidth}{!}{
    \begin{tabular}{c|c|ccccccccccccccccccc|c}
    \hline
    
    \hline
        \multicolumn{22}{c}{GTA5+SYNTHIA$\rightarrow$Cityscapes}\\
        \hline
        Setting & Method&\rotatebox{90}{road}&\rotatebox{90}{sidewalk}&\rotatebox{90}{building}&\rotatebox{90}{wall}&\rotatebox{90}{fence}&\rotatebox{90}{pole}&\rotatebox{90}{traffic light}&\rotatebox{90}{traffic sign}&\rotatebox{90}{vegetation}&\rotatebox{90}{terrian}&\rotatebox{90}{sky}&\rotatebox{90}{person}&\rotatebox{90}{rider}&\rotatebox{90}{car}&\rotatebox{90}{truck}&\rotatebox{90}{bus}&\rotatebox{90}{train}&\rotatebox{90}{motorbike}&\rotatebox{90}{bicycle}&mIoU\\
        \hline
        \multirow{8}{*}{NT} &
        Source & 3.0 & 10.1 & 42.1 & 7.3 & 6.6 & 10.6 & 18.2 & \textbf{31.2} & 61.3 & 3.9 & 73.2 & 27.5 & 16.2 & 9.9 & 1.4 & 1.6 & 0.0 & 8.6 & 3.8 & 17.7\\
        & AdaptSegNet\cite{Tsai_adaptseg_2018} & 0.1 & 0.0 & 1.4 & 4.0 & 6.6 & 5.4 & 14.6 & 22.8 & 5.9 & 1.9 & 35.9 & 1.3 & 18.0 & 0.6 & 3.0 & 1.8 & 0.7 & 13.0 & 9.4 & 7.7\\
        & MinEnt\cite{vu2018advent} & 32.0 & 10.0 & 73.0 & 15.4 & 18.1 & 20.5 & 29.5 & 19.9 & 75.3 & 3.9 & \textbf{79.6} & 51.3 & 18.7 & 18.5 & 4.3 & 4.8 & \textbf{9.2} & \textbf{20.3} & 10.3 & 27.1\\
        & Advent\cite{vu2018advent} & 6.3 & 1.0 & 27.7 & 4.5 & 6.3 & 6.5 & 16.9 & 19.3 & 16.7 & 2.0 & 40.6 & 6.8 & 17.1 & 7.7 & 3.7 & 6.6 & 1.2 & 15.0 & 18.5 & 11.8\\
        & Ours (w/o PSF) & \textbf{82.8} & 30.8 & 78.9 & 17.5 & 15.8 & 28.0 & 34.8 & 18.9 & 79.1 & 10.5 & 78.4 & 52.0 & 18.2 & 71.4 & 16.8 & 34.3 & 2.0 & 11.0 & 8.0 & 36.3\\
        \cline{2-22}
        & Ours (ADV) & 82.1 & \textbf{35.2} & 78.1 & \textbf{27.3} & 18.8 & \textbf{29.6} & 33.0 & 21.1 & 78.3 & \textbf{36.9} & 75.3 & \textbf{58.9} & \textbf{25.0} & 69.6 & 19.3 & 33.8 & 0.0 & 15.6 & \textbf{22.9} & 40.1\\
        & Ours (PSF) & 77.8 & 31.9 & \textbf{79.5} & 17.9 & 18.1 & 29.0 & \textbf{34.9} & 20.9 & 80.2 & 9.0 & \textbf{79.6} & 55.6 & 20.9 & \textbf{74.4} & 16.9 & 25.5 & 0.0 & 17.0 & 18.9 & 37.3\\
        & Ours (ADV+PSF) & 81.7 & 34.1 & \textbf{79.5} & 26.7 & \textbf{19.4} & 29.0 & 32.0 & 23.2 & \textbf{82.3} & 31.4 & 79.5 & 57.5 & 22.3 & 66.6 & \textbf{26.8} & \textbf{40.2} & 0.0 & 19.4 & 20.4 & \textbf{40.6}\\
        \hline
        \multirow{8}{*}{T} &
        Source & 28.7 & 9.5 & 52.3 & 11.1 & 10.0 & 9.5 & 16.4 & \textbf{30.6} & 55.9 & 2.7 & 67.5 & 40.8 & 21.1 & 38.7 & 6.9 & 4.3 & \textbf{6.4} & \textbf{22.1} & 20.6 & 24.0\\
        &AdaptSegNet\cite{Tsai_adaptseg_2018} & 78.3 & 34.5 & 75.7 & 16.2 & 15.6 & 11.5 & 19.0 & 10.8 & 78.0 & 16.5 & 76.3 & 42.6 & 8.4 & 59.6 & 10.9 & 8.8 & 0.5 & 14.2 & 8.7 & 30.8\\
        &MinEnt\cite{vu2018advent} & 58.5 & 20.6 & 70.5 & 12.0 & 17.9 & 18.3 & 19.9 & 27.1 & 74.3 & 8.0 &79.1 & 46.5 & 20.5 & 37.7 & 9.1 & 20.4 & 2.8 & 18.9 & 10.6 & 30.1\\
        &Advent\cite{vu2018advent} & 78.0 & 34.3 & 75.9 & 14.5 & 5.8 & 9.8 & 17.2 & 10.2 & 76.4 & 15.0 & 76.9 & 40.6 & 3.1 & 61.3 & 19.3 & 14.5 & 0.0 & 9.9 & 12.5 & 30.3\\
        &Ours(w/o PSF) & 86.0 & 40.8 & 79.1 & 13.2 & 22.7 & 33.5 & 33.3 & 18.9 & 79.9 & 33.2 & 72.0 & 49.7 & 19.1 & 63.3 & 20.6 & 10.1 & 0.0 & 13.4 & 34.0 & 38.1\\
        \cline{2-22}
        &Ours (ADV) & 86.2 & 41.3 & 81.6 & 21.1 & \textbf{23.3} & 33.4 & 32.0 & 20.6 & 81.0 & 32.1 & \textbf{79.8} & 57.5 & \textbf{26.4} & 70.5 & \textbf{24.8} & \textbf{31.4} & 0.2 & 18.3 & 27.1 & 41.5\\
        &Ours (PSF) & \textbf{87.8} & \textbf{42.9} & 81.2 & 17.3 & 22.0 & 34.1 & \textbf{36.9} & 17.9 & 82.2 & 34.2 & 73.6 & \textbf{58.9} & 25.1 & \textbf{76.5} & 24.4 & 28.9 & 0.1 & 19.8 & \textbf{41.9} & 42.4\\
        &Ours (PSF+ADV) & 86.8 & 42.5 & \textbf{82.5} & \textbf{23.0} & 23.1 & \textbf{34.4} & 36.3 & 29.1 & \textbf{82.9} & \textbf{34.3} & 76.5 & 56.5 & 24.1 & 75.5 & 23.6 & 17.3 & 0.3 & 22.0 & 41.6 & \textbf{42.8}\\
        \hline
        
        \hline
        
    \end{tabular}
    }
    \caption{Per-Class IoU on the mDALU 2D semantic image segmentation benchmark. ``NT" means source domain images are not translated with CycleGAN, and ``T" means source domain images are translated with CycleGAN. The mIoU results are reported over 19 classes. The best results are denoted in bold.}
    \label{tab:seg_2d_per_class}
\end{table*}

\begin{table*}
    \centering
    
    \begin{tabular}{c|c|cccccccccc|c}
    \hline
    
    \hline
        \multicolumn{13}{c}{Cityscapes+Nuscenes$\rightarrow$A2D2}\\
        \hline
        Modality &
        Method&\rotatebox{90}{road}&\rotatebox{90}{sidewalk}&\rotatebox{90}{building}&\rotatebox{90}{pole}&\rotatebox{90}{sign}&\rotatebox{90}{nature}&\rotatebox{90}{person}&\rotatebox{90}{car}&\rotatebox{90}{truck}&\rotatebox{90}{bike}&mIoU\\
        \hline
        \multirow{8}{*}{2D} & 
        Sources & 83.1 & 48.7 & 85.0 & \textbf{34.8} & 36.1 & 87.0 & 0.0 & 0.0 & 0.0 & 0.0 & 37.5\\
        &xMUDA & 68.2 & 13.8 & 22.3 & 22.1 & 15.7 & 3.6 & 0.1 & 15.9 & 1.6 & 0.0 & 16.3\\
        &ES + MinEnt & 55.7 & 15.6 & 64.9 & 19.8 & 21.7 & 45.2 & 0.0 & 0.0 & 0.0 & 0.0 & 22.3\\
        &ES + KL & 14.4 & 20.0 & 74.2 & 15.3 & 36.6 & 46.2 & 0.0 & 9.3 & 1.5 & 0.0 & 21.7 \\
        &xMUDA + AKL & 44.8 & 29.7 & 46.5 & 36.2 & 33.6 & 61.4 & 0.0 & 21.0 & 1.8 & 0.0 & 27.5\\
        &xMUDA + AKL + COMP & 70.3 & 38.1 & 76.4 & 25.0 & 30.5 & 80.8 & 0.0 & 0.0 & 0.0 & 0.0 & 32.1\\
        &Ours (w/o PSF) & 85.8 & 54.3 & 81.8 & 34.1 & 40.8 & 81.4 & 0.0 & 0.0 & 2.8 & 0.0 & 38.1\\
        \cline{2-13}
        &Ours & \textbf{92.8} & \textbf{59.9} & \textbf{90.0} & 30.4 & \textbf{60.7} & \textbf{90.6} & \textbf{13.8} & \textbf{71.6} & \textbf{39.1} & \textbf{0.4} & \textbf{54.9} \\
        \hline
        \multirow{8}{*}{3D} & 
        Source & 0.0 & 0.0 & 0.0 & 0.0 & 0.0 & 0.0 & 2.1 & 16.1 & 1.4 & 0.0 & 2.0\\
        &xMUDA & 0.0 & 0.0 & 0.0 & 0.0 & 0.0 & 0.0 & 1.2 & 14.6 & 1.5 & 0.0 & 1.7\\
        &ES + MinEnt & 0.0 & 0.0 & 0.0 & 0.0 & 0.0 & 0.0 & 1.9 & 12.0 & 1.5 & 0.0 & 1.5\\
        &ES + KL & 0.0 & 0.0 & 0.0 & 0.0 & 0.0 & 0.0 & 1.9 & 9.8 & 1.8 & 1.2 & 1.5\\
        &xMUDA + AKL & 0.0 & 0.0 & 0.0 & 0.0 & 0.0 & 0.0 & 2.4 & 18.5 & 1.3 & 0.4 & 2.3\\
        &xMUDA + AKL + COMP & 6.1 & 1.8 & 0.0 & 0.0 & 0.0 & 0.0 & 2.1 & 17.9 & 1.3 & 0.0 & 2.9 \\
        &Ours (w/o PSF) & 0.6 & 0.7 & 0.3 & 0.0 & 0.0 & 2.4 & 1.3 & 16.1 & 2.3 & 0.0 & 2.4\\
        \cline{2-13}
        &Ours & \textbf{82.0} & \textbf{27.7} & \textbf{80.3} & \textbf{1.4} & \textbf{7.5} & \textbf{80.8} & \textbf{7.2} & \textbf{54.9} & \textbf{25.6} & \textbf{3.5} & \textbf{37.1}\\
        \hline
        \multirow{8}{*}{Fuse} & 
        Source & 85.5 & 51.8 & 83.8 & \textbf{41.8} & 40.2 & 83.8 & 6.3 & 23.0 & 8.8 & 0.0 & 42.5\\
        &xMUDA & 55.8 & 2.2 & 2.8 & 3.5 & 2.8 & 0.2 & 2.7 & 19.4 & 1.7 & 0.0 & 9.1\\
        & ES + MinEnt & 63.1 & 7.5 & 69.7 & 9.0 & 13.8 & 30.2 & 2.6 & 11.0 & 1.2 & 0.0 & 20.8\\
        &ES + KL & 10.6 & 21.2 & 65.0 & 18.2 & 26.8 & 34.7 & 5.4 & 12.0 & 2.4 & 0.3 & 19.7\\
        &xMUDA + AKL & 13.2 & 36.9 & 20.1 & 34.1 & 31.1 & 44.5 & 4.6 & 24.8 & 1.7 & 0.1 & 21.1\\
        &xMUDA + AKL + COMP & 74.1 & 43.5 & 74.4 & 35.2 & 35.5 & 71.0 & 4.1 & 34.7 & 5.0 & 0.0 & 37.7\\
        &Ours(w/o PSF) & 91.1 & 57.3 & 85.7 & 39.7 & 47.4 & 85.9 & 8.6 & 57.8 & 25.3 & 0.4 & 49.9\\
        \cline{2-13}
        &Ours & \textbf{91.7} & \textbf{58.6} & \textbf{90.1} & 34.5 & \textbf{58.8} & \textbf{90.3} & \textbf{15.4} & \textbf{72.4} & \textbf{43.6} & \textbf{1.3} & \textbf{55.7}\\
        
        \hline
        
        \hline
        
    \end{tabular}
    
    \caption{Per-Class IoU on the mDALU cross-modal semantic segmentation benchmark. The mIoU results are reported over 10 classes. The best results are denoted in bold.}
    \label{tab:cross_perclass}
\end{table*}

\begin{figure*}
    \centering
    \includegraphics[width=\textwidth]{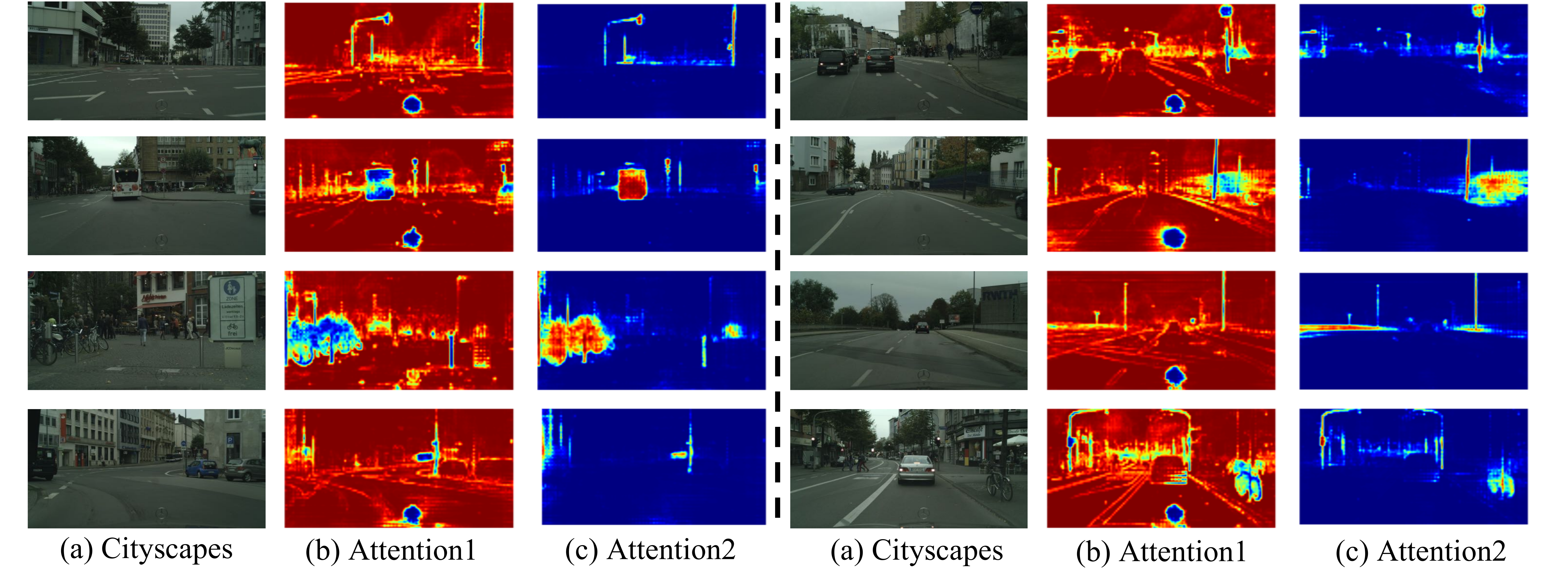}
    \caption{Visualization of the attention map $\tilde{\a}_{1}^{t}$ and $\tilde{\a}_{2}^{t}$ of the target domain images. (a) is the Cityscapes image $\x^{t}$. (b) is the attention map $\tilde{\a}_{1}^{t}$, generated by feeding the $\x^{t}$ into the attention network $M_{1}$. (c) is the attention map $\tilde{\a}_{2}^{t}$, generated by feeding the $\x^{t}$ into the attention network $M_{2}$. Red parts are the parts with higher attention value, while the blue parts with lower attention value.}
    \label{fig:atten_vis}
\end{figure*}

\textbf{Attention visualization for semantic segmentation.} During the ``partially-supervised adaptation" stage, we introduce the attention map in the domain attention (DAT) module, the attention-guided adversarial alignment (A$^3$) module and the inference via attention-guided fusion. In order to verify the effectiveness of our attention map prediction, we show the qualitative visualization of the attention map on the target domain images in Fig. \ref{fig:atten_vis}. Corresponding to the Sec. \textcolor{red}{3.2.1} of the main paper, the attention map $\tilde{\a}_{1}^{t}$ and $\tilde{\a}_{2}^{t}$, are generated by feeding the target domain image $\x^t$ into the attention network $M_1$ and $M_2$. It is shown that our predicted attention map $\tilde{\a}_{1}^{t}$, corresponding to the source domain $\cS_1$, has higher attention value, for the objects belonging to the partial label space $\cC_1$, such as the road, sidewalk, building, vegetation, sky and car. And the predicted attention map $\tilde{\a}_{2}^{t}$, corresponding to the source domain $\cS_2$, has higher attention value, for the objects belonging to the partial label space $\cC_2$, such as
the fence, pole, light, sign, bus, motorcycle and bicycle. It proves the validity of our attention map prediction.

\begin{figure*}
    \centering
    \includegraphics[width=\linewidth]{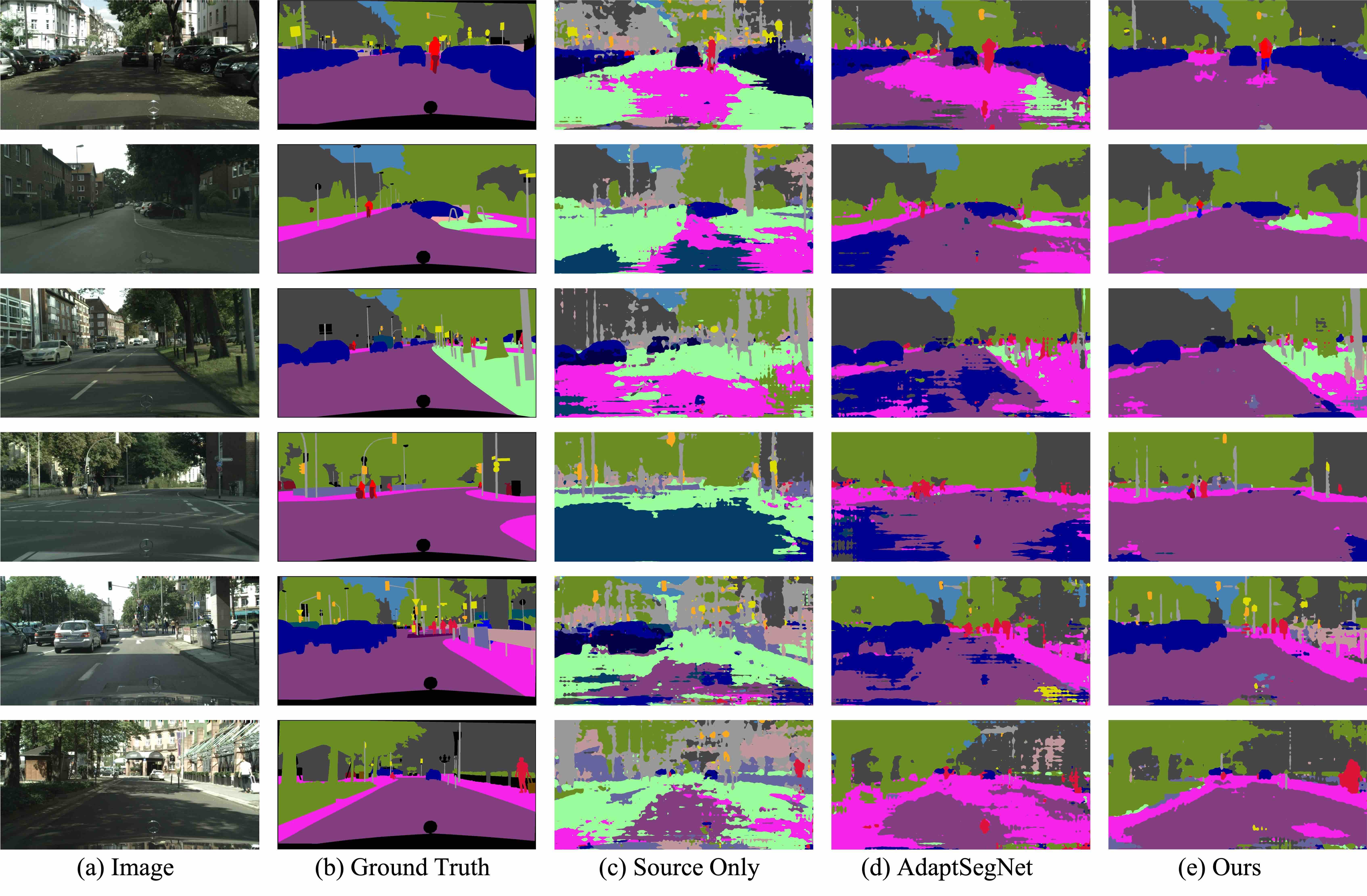}
    \caption{Qualitative comparison of semantic segmentation results, under the mDALU 2D semantic image segmentation benchmark, SYNTHIA + GTA5 $\rightarrow$ Cityscapes. The source images are translated with CycleGAN, \ie setting ``T".}
    \label{fig:seg_vis_2d_supp}
\end{figure*}

\textbf{Additional qualitative results for semantic segmentation.} In Fig. \textcolor{red}{4} of the main paper, we show the qualitative comparison results between our approach and other methods on the 2D semantic image segmentation benchmark, and the source domain images are not translated with CycleGAN~\cite{CycleGAN2017}, \ie the ``NT" setting. Here we provide additional qualitative comparison results between our approach and other methods on the 2D semantic image segmentation benchmark, and the source images are translated with CycleGAN~\cite{CycleGAN2017}, \ie the ``T" setting. As shown in Fig. \ref{fig:seg_vis_2d_supp}, it can be seen that our approach obviously outperforms other methods on the 2D semantic image segmentation benchmark. It further verifies the effectiveness of our approach to mDALU problem.

\textbf{Comparison between w/ and w/o relabeling inconsistent taxonomies in the source domain.} In Sec. \textcolor{red}{3.2.6} of the main paper, we introduce the extension of our method for inconsistent taxonomies. In the PSF module, besides the unlabeled samples in the source domain being completed with the predicted pseudo-label as in Eq. (\textcolor{red}{12}), we add Eq.~(\textcolor{red}{17}) to relabel the conflict part $\c_{p}^{q} \cap \c_{m}^{n}$ in the source domain $\cS_m$. In Table~\textcolor{red}{7} of the main paper, we show the performance of our extended method $40.0\%$ under inconsistent taxonomies setting, which outperforms other competing methods significantly and proves the effectiveness of our extended method for inconsistent taxonomies. Here, we compare the ablation of our extended method, w/o relabeling inconsistent taxonomies in the source domain, against our full extended method, to further verify the effectiveness of relabeling inconsistent taxonomies as in Eq. (\textcolor{red}{17}). From Table~\ref{tab:seg_icon_relabel}, it is shown that the performance is improved by $1.6\%$ from $38.4\%$ to $40.0\%$, by relabeling inconsistent taxonomies in the source domain. And the detailed performance comparison on the inconsistent taxonomies classes in Table~\ref{tab:seg_icon_relabel} also proves the effectiveness of relabeling inconsistent taxonomies in the source domain.

\end{document}